\documentclass[sigconf]{acmart}

\copyrightyear{2020}
\acmYear{2020}
\setcopyright{acmcopyright}
\acmConference[ASE '20]{35th IEEE/ACM International Conference on Automated
Software Engineering}{September 21--25, 2020}{Virtual Event, Australia}
\acmBooktitle{35th IEEE/ACM International Conference on Automated Software
Engineering (ASE '20), September 21--25, 2020, Virtual Event, Australia}
\acmPrice{15.00}
\acmDOI{10.1145/3324884.3416560}
\acmISBN{978-1-4503-6768-4/20/09}

\usepackage{tikz}
\usepackage{graphicx}
\usepackage{pgf}

\usepackage{diagbox}
\usepackage{algorithm2e}
\usepackage{makecell}
\usepackage{subcaption}
\usepackage{setspace}
\usepackage{multirow}
\usepackage{amsmath}

\definecolor{mygreen}{rgb}{0,0.6,0}
\definecolor{mygray}{rgb}{0.5,0.5,0.5}
\definecolor{mymauve}{rgb}{0.58,0,0.82}
\definecolor{dkgreen}{rgb}{0,0.5,0}

\definecolor{lightgray}{rgb}{0.85,0.85,0.85}
\definecolor{lightgreen}{rgb}{0.7,0.9,0.7}
\definecolor{lightblue}{rgb}{0.7,0.7,0.9}
\definecolor{lightred}{rgb}{0.9,0.7,0.7}

\newcommand{\ignore}[1]{}

\newcommand{\Name}{\textsc{NeuroDiff}}
\newcommand{\ReluDiffP}{\textsc{ReluDiff+}}
\newcommand{\ReluDiff}{\textsc{ReluDiff}}

\newcommand{\ReluVal}{\textsc{ReluVal}}
\newcommand{\DeepPoly}{\textsc{DeepPoly}}

\newcommand{\Neurify}{\textsc{Neurify}}

\newcommand{\nj}[0]{n_{k,j}}
\newcommand{\n}[2]{n_{#1,#2}}
\newcommand{\np}[2]{n'_{#1,#2}}
\newcommand{\npj}[0]{n'_{k,j}}
\newcommand{\nd}[2]{\Delta_{#1,#2}}
\newcommand{\ndj}{\Delta_{k,j}}
\newcommand{\W}[3]{W_{#1}[#2,#3]}
\newcommand{\Wp}[3]{W_{#1}'[#2,#3]}
\newcommand{\Wk}[1]{W_{#1}}
\newcommand{\Wd}[3]{W_{#1}^\Delta[#2,#3]}

\newcommand{\IntIn}[1]{S^{in}(#1)}
\newcommand{\IntOut}[1]{S(#1)}

\newcommand{\relu}[1]{ReLU(#1)}

\newcommand{\UBEq}[1]{\mathsf{UB}(#1)}
\newcommand{\LBEq}[1]{\mathsf{LB}(#1)}
\newcommand{\UBU}[1]{\overline{\mathsf{UB}}(#1)}
\newcommand{\UBL}[1]{\underline{\mathsf{UB}}(#1)}
\newcommand{\LBL}[1]{\underline{\mathsf{LB}}(#1)}
\newcommand{\LBU}[1]{\overline{\mathsf{LB}}(#1)}

\theoremstyle{plain}

\newtheoremstyle{case}{}{}{}{}{}{:}{ }{}
\theoremstyle{case}

\begin{document}

\title[]{\Name{}: Scalable Differential Verification of Neural Networks using Fine-Grained Approximation}

\author{Brandon Paulsen}
\affiliation{
  \institution{University of Southern California}
  \city{Los Angeles}
  \state{California}
  \country{USA}
}
\author{Jingbo Wang}
\affiliation{
  \institution{University of Southern California}
  \city{Los Angeles}
  \state{California}
  \country{USA}
}
\author{Jiawei Wang}
\affiliation{
	\institution{University of Southern California}
	\city{Los Angeles}
	\state{California}
	\country{USA}
}
\author{Chao Wang}
\affiliation{
  \institution{University of Southern California}
  \city{Los Angeles}
  \state{California}
  \country{USA}
}

\begin{abstract}
As neural networks make their way into safety-critical systems, where
misbehavior can lead to catastrophes, there is a growing interest in
certifying the equivalence of two structurally similar neural
networks -- a problem known as \textit{differential verification}.
For example, compression techniques are often used in practice for
deploying trained neural networks on computationally- and
energy-constrained devices, which raises the question of how
faithfully the compressed network mimics the original network.
Unfortunately, existing methods either focus on verifying a single
network or rely on loose approximations to prove the equivalence of
two networks. Due to overly conservative approximation, differential
verification lacks scalability in terms of both accuracy and
computational cost.
To overcome these problems, we propose \Name{}, a \emph{symbolic}
and \emph{fine-grained} approximation technique that drastically
increases the accuracy of differential verification on feed-forward ReLU
networks while achieving many orders-of-magnitude speedup.
\Name{} has two key contributions.  The first one is new convex
approximations that more accurately bound the difference of two
networks under all possible inputs.  The second one is judicious use of
symbolic variables to represent neurons whose difference bounds have
accumulated significant error.  We find that these two techniques are
complementary, i.e., when combined, the benefit is greater than the
sum of their individual benefits.
We have evaluated \Name{} on a variety of differential verification
tasks. Our results show that \Name{} is up to 1000X faster and 5X more
accurate than the state-of-the-art tool.
\end{abstract}

\maketitle

\section{Introduction}
\label{sec:introduction}

%\subsubsection{Need for differential verification}
There is a growing need for rigorous analysis techniques that can
compare the behaviors of two or more neural networks trained for the
same task. For example, such techniques have applications in
better understanding the representations learned by different
networks~\cite{wang2018towards}, and finding inputs where networks
disagree~\cite{xie2019diffchaser}.
The need is further motivated by the increasing use of neural network
compression~\cite{HanMD16} -- a technique that alters the network's
parameters to reduce its energy and computational cost -- where
we \textit{expect} the compressed network to be functionally
equivalent to the original network.
In safety-critical systems where a single instance of misbehavior can
lead to catastrophe, having \textit{formal guarantees} on the
equivalence of the original and compressed networks is
highly desirable.

%\subsubsection{Limitations of existing methods}
Unfortunately, most work aimed at verifying or testing neural networks
does not provide formal guarantees on their equivalence.  For example,
testing techniques geared toward \emph{refutation} can provide inputs
where a single network misbehaves~\cite{ma2018deepgauge,
xie2019deephunter, SunWRHKK18, TianPJR18, odena2018tensorfuzz} or
multiple networks disagree~\cite{xie2019diffchaser,PeiCYJ17,MaLLZG18},
but they do not guarantee the absence of misbehaviors or disagreements.
While techniques geared toward \emph{verification} can prove safety or
robustness properties of a single
network~\cite{HuangKWW17,Ehlers17,KatzHIJLLSTWZDK19,RuanHK18,
WangPWYJ18nips,SinghGPV19iclr,MirmanGV18,GehrMDTCV18,FischerBDGZV19},
they lack crucial information needed to prove the equivalence of
multiple networks.
One exception is the \ReluDiff{} tool of Paulsen et
al.~\cite{PaulsenWW20}, which computes a sound approximation of the
difference of two neural networks, a problem known as
\textit{differential verification}.  While \ReluDiff{} performs
better than other techniques, the overly conservative approximation it
computes often causes both accuracy and efficiency to suffer.

%\subsubsection{Our main contributions}
To overcome these problems, we propose \Name{}, a new \emph{symbolic}
and \emph{fine-grained} approximation technique that significantly increases the
accuracy of differential verification while achieving many
orders-of-magnitude speedup.
\Name{} has two key contributions.  The first contribution is the development
of \emph{convex approximations}, a fine-grained approximation technique
for bounding the output difference of neurons for all possible inputs,
which drastically improves over the
coarse-grained \emph{concretizations} used by \ReluDiff{}.
The second contribution is judiciously introducing symbolic variables
to represent neurons in hidden layers whose difference bounds have
accumulated significant approximation error.
These two techniques are also complementary, i.e., when combined, the
benefit is significantly greater than the sum of their individual
benefits.

\begin{figure}[t]
\scalebox{0.85}{\centering \input{figs/diagram.tex}}
\caption{The overall flow of \Name{}.}
\label{fig:diagram}
\end{figure}

%\subsubsection{Overall flow of our method}
The overall flow of \Name{} is shown in Figure~\ref{fig:diagram},
where it takes as input two neural networks $ f $ and $ f' $, a set of
inputs to the neural networks $ X $ defined by box intervals, and a small
constant $ \epsilon $ that quantifies the tolerance for disagreement. We assume
that $ f $ and $ f' $ have the same network topology and only differ in the
numerical values of their weights. In practice, $ f' $ could be the compressed
version of $ f $, or they could be networks constructed using the same network
topology but slightly different training data. We also note that this
assumption can support compression techniques such as weight
pruning~\cite{HanMD16} (by setting edges' weights to 0) and even neuron
removal~\cite{gokulanathan2019simplifying} (by setting all of a neuron's
incoming edge weights to 0). \Name{} then aims to prove $
\forall
x \in X. |f'(x) - f(x)| < \epsilon $. It can return
(1) \emph{verified} if a proof can be found, or
(2) \emph{undetermined} if a specified timeout is reached.

Internally, \Name{} first performs a forward analysis using symbolic
interval arithmetic to bound both the absolute value ranges of all
neurons, as in single network verification, and the difference between
the neurons of the two networks. \Name{} then checks if the difference
between the output neurons satisfies $ \epsilon $, and if so
returns \emph{verified}. Otherwise,
\Name{} uses a gradient-based refinement to partition $ X
$ into two disjoint sub regions $ X_1 $ and $ X_2 $, and attempts the
analysis again on the individual regions. Since $ X_1 $ and $ X_2 $
form independent sub-problems, we can do these analyses in parallel,
hence gaining significant speedup.
%  (a) interval arithmetic to propagate absolute value ranges
%  (b) interval arithmetic to propagate difference value ranges
%  (c) multi-threading to perform forward interval analysis
%  (d) gradient based refinement to partition input regions

%\subsubsection{Convex approximation: what's new?}
The new convex approximations used in \Name{} are significantly more
accurate than not only the coarse-grained \emph{concretizations}
in \ReluDiff{}~\cite{PaulsenWW20} but also the standard convex
approximations in single-network verification tools~\cite{SinghGPV19,
Singh2019krelu, WangPWYJ18nips, zhang2018efficient}.
While these (standard) convex approximations aim to bound the absolute
value range of $ y = \relu{x} $, where $x$ is the input of
the \textit{rectified linear unit} (ReLU) activation function, our new
convex approximations aim to bound the difference $ z = \relu{x
+ \Delta} - \relu{x} $, where $x$ and $x+\Delta$ are ReLU inputs of
two corresponding neurons.
This is significantly more challenging because it involves the search
of bounding planes in a three-dimensional space (defined by $x$,
$\Delta$ and $z$) as opposed to a two-dimensional space as in the
prior work.

%\subsubsection{Symbolic variables: what's new?}
The symbolic variables we judiciously add to represent values of
neurons in hidden layers should not be confused with the symbolic
inputs used by existing tools either.
While the use of symbolic inputs is well understood, e.g., both in
single-network verification~\cite{SinghGPV19, Singh2019krelu,
WangPWYJ18nips, zhang2018efficient} and differential
verification~\cite{PaulsenWW20}, this is the first time that symbolic
variables are used to substitute values of hidden neurons during
differential verification.
While the impact of symbolic inputs often diminishes after the first
few layers of neurons, the impact of these new symbolic variables,
when judiciously added, can be maintained in any hidden layer.

%\subsubsection{Experiments: does it work?}
We have implemented the proposed \Name{} in a tool and evaluated it on
a large set of differential verification tasks. Our benchmarks
consists of 49 networks, from applications such as aircraft collision
avoidance, image classification, and human activity recognition.  We
have experimentally compared with \ReluDiff{}~\cite{PaulsenWW20}, the
state-of-the-art tool which has also been shown to be superior to
\ReluVal{}~\cite{WangPWYJ18} and \DeepPoly{}~\cite{SinghGPV19} for differential
verification.
Our results show that \Name{} is up to 1,000X faster and 5X more
accurate.  In addition, \Name{} is able to prove many of the same properties
as \ReluDiff{} while considering much larger input regions.

To summarize, this paper makes the following contributions:
\begin{itemize}
\item
We propose new convex approximations to more accurately bound the
difference between corresponding neurons of two structurally similar
neural networks.
\item
We propose a method for judiciously introducing symbolic variables to
neurons in hidden layers to mitigate the propagation of approximation
error.
\item
We implement and evaluate the proposed technique on a large number of
differential verification tasks and demonstrate its significant speed
and accuracy gains.
\end{itemize}

The remainder of this paper is organized as follows.  First, we
provide a brief overview of our method in Section~\ref{sec:overview}.
Then, we provide the technical background in
Section~\ref{sec:preliminary}.  Next, we present the detailed
algorithms in Section~\ref{sec:approach} and the experimental results
in Section~\ref{sec:experiment}.  We review the related work in
Section~\ref{sec:related}. Finally, we give our conclusions in
Section~\ref{sec:conclusion}.

\section{Overview}
\label{sec:overview}

In this section, we highlight our main contributions and illustrate
the shortcomings of previous work on a motivating example.

\begin{figure}
\scalebox{.72}{\tikzset{every picture/.style={line width=0.75pt}} %set default line width to 0.75pt        

\begin{tikzpicture}[x=0.75pt,y=0.75pt,yscale=-1,xscale=1]
%uncomment if require: \path (0,395); %set diagram left start at 0, and has height of 395

%Shape: Circle [id:dp29452950296941915] 
%\draw   (90,55) .. controls (90,41.19) and (101.19,30) .. (115,30) .. controls (128.81,30) and (140,41.19) .. (140,55) .. controls (140,68.81) and (128.81,80) .. (115,80) .. controls (101.19,80) and (90,68.81) .. (90,55) -- cycle ;
\draw (115,55) +(-25,-20) rectangle +(25,20) ;
%Shape: Circle [id:dp7730288688046164] 
%\draw   (90,160) .. controls (90,146.19) and (101.19,135) .. (115,135) .. controls (128.81,135) and (140,146.19) .. (140,160) .. controls (140,173.81) and (128.81,185) .. (115,185) .. controls (101.19,185) and (90,173.81) .. (90,160) -- cycle ;
\draw (115,160) +(-25,-20) rectangle +(25,20) ;
%Shape: Circle [id:dp7848793538270885] 
\draw   (190,160) .. controls (190,146.19) and (201.19,135) .. (215,135) .. controls (228.81,135) and (240,146.19) .. (240,160) .. controls (240,173.81) and (228.81,185) .. (215,185) .. controls (201.19,185) and (190,173.81) .. (190,160) -- cycle ;
%Shape: Circle [id:dp2581761932666119] 
\draw   (190,55) .. controls (190,41.19) and (201.19,30) .. (215,30) .. controls (228.81,30) and (240,41.19) .. (240,55) .. controls (240,68.81) and (228.81,80) .. (215,80) .. controls (201.19,80) and (190,68.81) .. (190,55) -- cycle ;
%Shape: Circle [id:dp03162623697960232] 
\draw   (290,160) .. controls (290,146.19) and (301.19,135) .. (315,135) .. controls (328.81,135) and (340,146.19) .. (340,160) .. controls (340,173.81) and (328.81,185) .. (315,185) .. controls (301.19,185) and (290,173.81) .. (290,160) -- cycle ;
%Shape: Circle [id:dp02546394024001375] 
\draw   (290,55) .. controls (290,41.19) and (301.19,30) .. (315,30) .. controls (328.81,30) and (340,41.19) .. (340,55) .. controls (340,68.81) and (328.81,80) .. (315,80) .. controls (301.19,80) and (290,68.81) .. (290,55) -- cycle ;
%Shape: Circle [id:dp12084775666404268] 
%\draw   (371,105) .. controls (371,91.19) and (382.19,80) .. (396,80) .. controls (409.81,80) and (421,91.19) .. (421,105) .. controls (421,118.81) and (409.81,130) .. (396,130) .. controls (382.19,130) and (371,118.81) .. (371,105) -- cycle ;
\draw (396,105) +(-25,-20) rectangle +(25,20) ;
%Straight Lines [id:da34998381745971496] 
\draw    (140,160) -- (187,160) ;
\draw [shift={(190,160)}, rotate = 180] [fill={rgb, 255:red, 0; green, 0; blue, 0 }  ][line width=0.08]  [draw opacity=0] (10.72,-5.15) -- (0,0) -- (10.72,5.15) -- (7.12,0) -- cycle    ;
%Straight Lines [id:da4638805061238097] 
\draw    (140,160) -- (188.71,57.71) ;
\draw [shift={(190,55)}, rotate = 475.46] [fill={rgb, 255:red, 0; green, 0; blue, 0 }  ][line width=0.08]  [draw opacity=0] (10.72,-5.15) -- (0,0) -- (10.72,5.15) -- (7.12,0) -- cycle    ;
%Straight Lines [id:da15742477491714235] 
\draw    (140,55) -- (187,55) ;
\draw [shift={(190,55)}, rotate = 180] [fill={rgb, 255:red, 0; green, 0; blue, 0 }  ][line width=0.08]  [draw opacity=0] (10.72,-5.15) -- (0,0) -- (10.72,5.15) -- (7.12,0) -- cycle    ;
%Straight Lines [id:da9054930639679811] 
\draw    (140,55) -- (188.71,157.29) ;
\draw [shift={(190,160)}, rotate = 244.54000000000002] [fill={rgb, 255:red, 0; green, 0; blue, 0 }  ][line width=0.08]  [draw opacity=0] (10.72,-5.15) -- (0,0) -- (10.72,5.15) -- (7.12,0) -- cycle    ;
%Straight Lines [id:da0727882271401632] 
\draw    (240,160) -- (288.71,57.71) ;
\draw [shift={(290,55)}, rotate = 475.46] [fill={rgb, 255:red, 0; green, 0; blue, 0 }  ][line width=0.08]  [draw opacity=0] (10.72,-5.15) -- (0,0) -- (10.72,5.15) -- (7.12,0) -- cycle    ;
%Straight Lines [id:da02068180741540726] 
\draw    (240,55) -- (288.71,157.29) ;
\draw [shift={(290,160)}, rotate = 244.54000000000002] [fill={rgb, 255:red, 0; green, 0; blue, 0 }  ][line width=0.08]  [draw opacity=0] (10.72,-5.15) -- (0,0) -- (10.72,5.15) -- (7.12,0) -- cycle    ;
%Straight Lines [id:da4941096166166473] 
\draw    (340,55) -- (369.42,102.45) ;
\draw [shift={(371,105)}, rotate = 238.2] [fill={rgb, 255:red, 0; green, 0; blue, 0 }  ][line width=0.08]  [draw opacity=0] (10.72,-5.15) -- (0,0) -- (10.72,5.15) -- (7.12,0) -- cycle    ;
%Straight Lines [id:da0005753242144755921] 
\draw    (340,160) -- (369.53,107.61) ;
\draw [shift={(371,105)}, rotate = 479.41] [fill={rgb, 255:red, 0; green, 0; blue, 0 }  ][line width=0.08]  [draw opacity=0] (10.72,-5.15) -- (0,0) -- (10.72,5.15) -- (7.12,0) -- cycle    ;
%Straight Lines [id:da6663993224468625] 
\draw    (240,160) -- (287,160) ;
\draw [shift={(290,160)}, rotate = 180] [fill={rgb, 255:red, 0; green, 0; blue, 0 }  ][line width=0.08]  [draw opacity=0] (10.72,-5.15) -- (0,0) -- (10.72,5.15) -- (7.12,0) -- cycle    ;
%Straight Lines [id:da971612493075633] 
\draw    (240,55) -- (287,55) ;
\draw [shift={(290,55)}, rotate = 180] [fill={rgb, 255:red, 0; green, 0; blue, 0 }  ][line width=0.08]  [draw opacity=0] (10.72,-5.15) -- (0,0) -- (10.72,5.15) -- (7.12,0) -- cycle    ;

% Text Node
\draw (165,43) node  [font=\normalsize] [align=left] {1.9};
% Text Node
\draw (140,90) node  [font=\normalsize] [align=left] {1.1};
% Text Node
\draw (140,120) node  [font=\normalsize] [align=left] {\mbox{-}1.9};
% Text Node
\draw (165,172.5) node  [font=\normalsize] [align=left] {1.0};
% Text Node
\draw (265,42.5) node  [font=\normalsize] [align=left] {2.1};
% Text Node
\draw (238,90) node  [font=\normalsize] [align=left] {0.9};
% Text Node
\draw (265,172.5) node  [font=\normalsize] [align=left] {1.1};
% Text Node
\draw (238,120) node  [font=\normalsize] [align=left] {\mbox{-}1.0};
% Text Node
\draw (366.5,69) node  [font=\normalsize] [align=left] {1.0};
% Text Node
\draw (369,141) node  [font=\normalsize] [align=left] {\mbox{-}1.0};
% Text Node
\draw (115,55) node  {$n_{0,1}$};
% Text Node
\draw (115,160) node    {$n_{0,2}$};
% Text Node
\draw (215,160) node    {$n_{1,2}$};
% Text Node
\draw (215,55) node    {$n_{1,1}$};
% Text Node
\draw (315,55) node    {$n_{2,1}$};
% Text Node
\draw (315,160) node    {$n_{2,2}$};
% Text Node
\draw (396,105) node    {$n_{3,1}$};
% Text Node
\draw (55,57) node    {$x_{1} \in [ -2,2]$};
% Text Node
\draw (54,161) node    {$x_{2} \in [ -2,2]$};
% Text Node
\draw (118.5,120) node  [font=\normalsize,color={rgb, 255:red, 74; green, 144; blue, 226 }  ,opacity=1 ] [align=left] {\mbox{-}2.0};
% Text Node
\draw (119.5,90) node  [font=\normalsize,color={rgb, 255:red, 74; green, 144; blue, 226 }  ,opacity=1 ] [align=left] {1.0};
% Text Node
\draw (165,29.5) node  [font=\normalsize,color={rgb, 255:red, 74; green, 144; blue, 226 }  ,opacity=1 ] [align=left] {2.0};
% Text Node
\draw (165,185.5) node  [font=\normalsize,color={rgb, 255:red, 74; green, 144; blue, 226 }  ,opacity=1 ] [align=left] {1.0};
% Text Node
\draw (213.5,120) node  [font=\normalsize,color={rgb, 255:red, 74; green, 144; blue, 226 }  ,opacity=1 ] [align=left] {\mbox{-}1.0};
% Text Node
\draw (213.5,90) node  [font=\normalsize,color={rgb, 255:red, 74; green, 144; blue, 226 }  ,opacity=1 ] [align=left] {1.0};
% Text Node
\draw (265,29.5) node  [font=\normalsize,color={rgb, 255:red, 74; green, 144; blue, 226 }  ,opacity=1 ] [align=left] {2.0};
% Text Node
\draw (265,185.5) node  [font=\normalsize,color={rgb, 255:red, 74; green, 144; blue, 226 }  ,opacity=1 ] [align=left] {1.0};
% Text Node
\draw (366.5,56.5) node  [font=\normalsize,color={rgb, 255:red, 74; green, 144; blue, 226 }  ,opacity=1 ] [align=left] {1.0};
% Text Node
\draw (369,154.5) node  [font=\normalsize,color={rgb, 255:red, 74; green, 144; blue, 226 }  ,opacity=1 ] [align=left] {\mbox{-}1.0};

\end{tikzpicture}}
\caption{Motivating example.}
\label{fig:motex}
\end{figure}
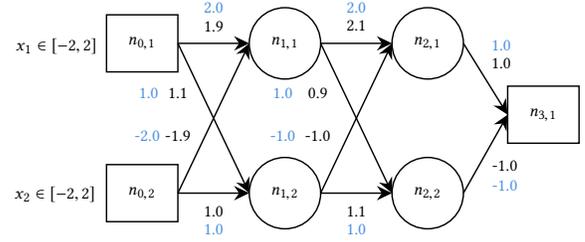

\subsection{Differential Verification}

We use the neural network in Figure~\ref{fig:motex} as a running
example. The network has two input nodes $ \n{0}{1}, \n{0}{2} $, two
hidden layers with two neurons each ($ \n{1}{1}, \n{1}{2} $ and
$ \n{2}{1}, \n{2}{2}$), and one output node $ \n{3}{1} $.  Each neuron
in the hidden layer performs a summation of their inputs, followed by
a \textit{rectified linear unit} (ReLU) activation function, defined
as $ y = max(0,x) $, where $ x $ is the input to the ReLU activation
function, and $ y $ is the output.

Let this entire network be $ f $, and the value of the output node be
$ \n{3}{1} = f(x_1, x_2) $, where $ x_1$ and $x_2 $ are the values of
input nodes $ \n{0}{1}$ and $\n{0}{2} $, respectively.
The network can be evaluated on a specific input by performing a
series matrix multiplications (i.e., affine transformations) followed
by element-wise ReLU transformations. For example, the output of
the neurons of the first hidden layer is
\[
\begin{bmatrix}
\n{1}{1} \\
\n{1}{2}
\end{bmatrix}
=
ReLU\Bigg(
\begin{bmatrix}
1.9 & -1.9 \\
1.0 & 1.1
\end{bmatrix}
\cdot
\begin{bmatrix}
x_1 \\
x_2
\end{bmatrix}
\Bigg)
=
\begin{bmatrix}
ReLU(1.9x_1 - 1.9x_2) \\
ReLU(1.1x_1 + 1.0x_2)
\end{bmatrix}
\]

Differential verification aims to compare $ f $ to another network $
f' $ that is structurally similar. For our example, $f'$ is
obtained by rounding the edge weights of $ f $ to the nearest whole
numbers, a network compression technique known as \textit{weight
quantization}.
Thus, $ f' $, $ \np{k}{j} $ and $ \np{3}{1} = f'(x_1, x_2) $ are
counterparts of $ f $, $ \n{k}{j} $ and $ \n{3}{1} = f(x_1, x_2) $ for
$0\leq k\leq 2$ and $1\leq j\leq 2$.
Our goal is to prove that $ |f'(x_1, x_2) - f(x_1, x_2)| $ is less
than some reasonably small $ \epsilon $ for all inputs defined by the
intervals $ x_1 \in [-2,2]$ and $x_2 \in [-2,2] $.
For ease of understanding, we show the edge weights of $ f $ in black, and
$ f' $ in light blue in Figure~\ref{fig:motex}.

\subsection{Limitations of Existing Methods}

Naively, one could adapt any state-of-the-art, single-network
verification tool for our task,
including \DeepPoly{}~\cite{SinghGPV19}
and \Neurify{}~\cite{WangPWYJ18nips}.
\Neurify{}, in particular, takes a neural network and
an input region of the network, and uses interval
arithmetic~\cite{moore2009introduction, WangPWYJ18} to produce sound
symbolic lower and upper bounds for each output node.
Typically, \Neurify{} would then use the computed bounds to certify
the absence of \textit{adversarial
examples}~\cite{szegedy2013intriguing} for the network.

However, for our task, the bounds must be computed for both networks $
f $ and $ f' $.  Then, we subtract them, and concretize to compute
lower and upper bounds on $ f'(x_1, x_2) - f(x_1, x_2) $.
In our example, the individual bounds would be (approximately, due to
rounding) $[LB(f),UB(f)] = [-0.94x_1 -0.62x_2 -6.51, $ $ 0.71x_1
-2.35x_2 +7.98] $ and $ [LB(f'),UB(f')] = [-0.94x_1 -0.44x_2 -6.75,$ $
0.75x_1 -2.25x_2 +8.00] $ for nodes $ \n{3}{1} $ and $ \np{3}{1} $,
respectively.  After the subtraction, we would obtain the bounds
$[LB(f')-UB(f), UB(f')-LB(f)] = $ $ [-1.65x_1 + 1.9x_2 - 14.73,
1.68x_1 - 1.63x_2 + 14.5] $.  After concretization, we would obtain
the bounds $ [-21.83, 21.12] $.  Unfortunately, the bounds are far
from being accurate.

The \ReluDiff{} method of Paulsen et al.~\cite{PaulsenWW20} showed
that, by directly computing a \textit{difference interval}
layer-by-layer, the accuracy can be greatly improved.  For the running
example, \ReluDiff{} would first compute bounds on the difference
between the neurons $ \n{1}{1} $ and $ \np{1}{1} $, which is $ [0,
1.1] $, and then similarly compute bounds on the difference between
outputs of $ \n{1}{2} $ and $ \np{1}{2} $. Then, the results would be
used to compute difference bounds of the subsequent layer.
The reason it is more accurate is because it begins computing part of
the difference bound \emph{before} errors have accumulated, whereas
the naive approach first accumulates significant errors at each
neuron, and \emph{then} computes the difference bound.
In our running example, \ReluDiff{}~\cite{PaulsenWW20} would
compute the tighter bounds $ [-3.1101, 2.5600] $.

While \ReluDiff{} improves over the naive approach, in many cases,
it uses \emph{concrete} values for the upper and lower bounds.
In practice, this approach can suffer from severe error-explosion.
Specifically,
whenever a neuron of either network is in an \textit{unstable} state --
i.e., when a ReLU's input interval contains the value 0 -- it has to
concretize the symbolic expressions.

\subsection{Our Method}

\begin{figure*}
\centering
\begin{minipage}[t]{0.45\linewidth}
\centering
%\captionsetup{width=0.98\textwidth}
\includegraphics[width=\linewidth]{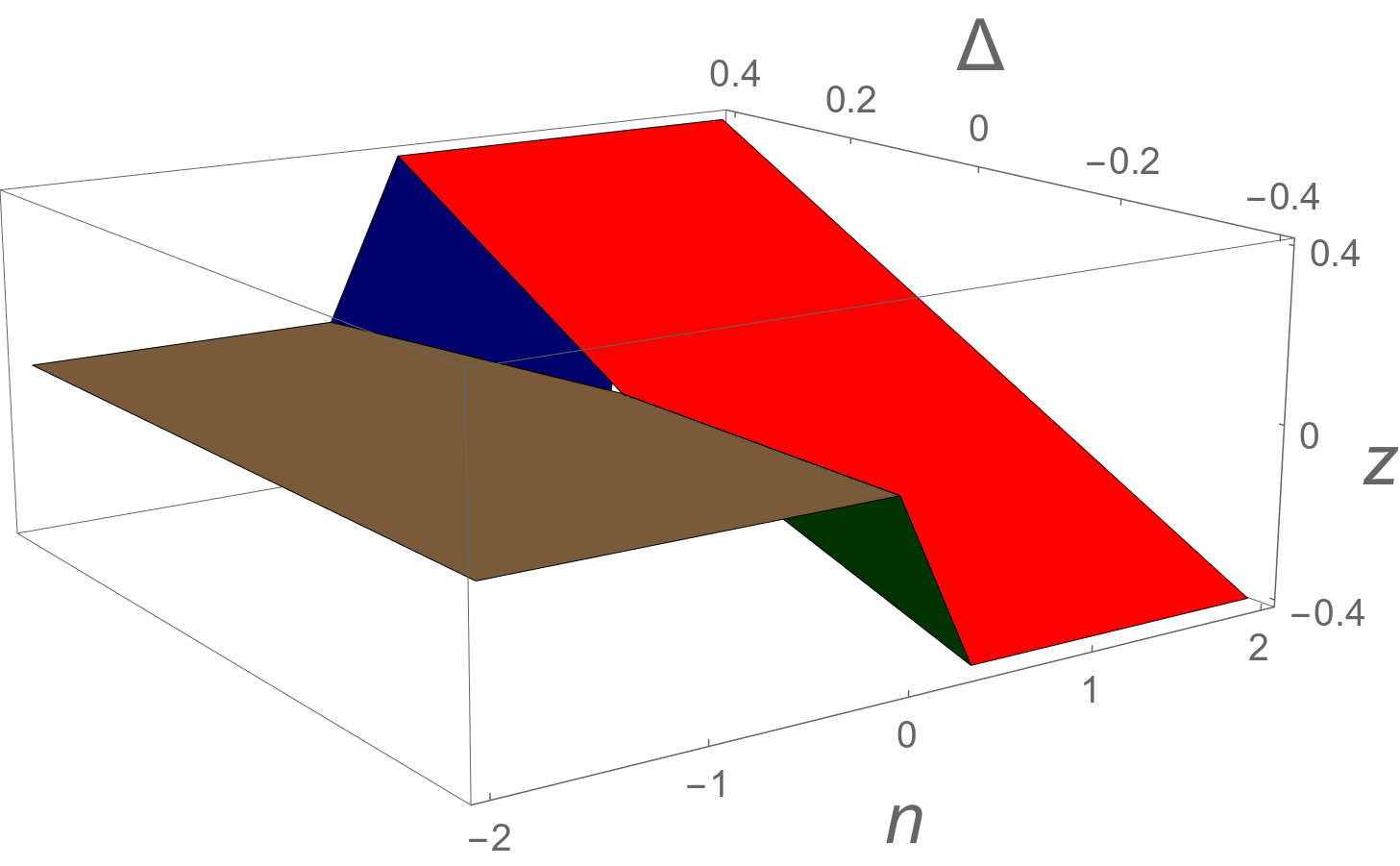}
\caption{The shape of $ z = ReLU(n + \Delta) - ReLU(n) $.}
\label{fig:equation1}
\end{minipage}
\hspace{0.09\linewidth}
\begin{minipage}[t]{0.45\linewidth}
\centering
%\captionsetup{width=0.98\textwidth}
\includegraphics[width=\linewidth]{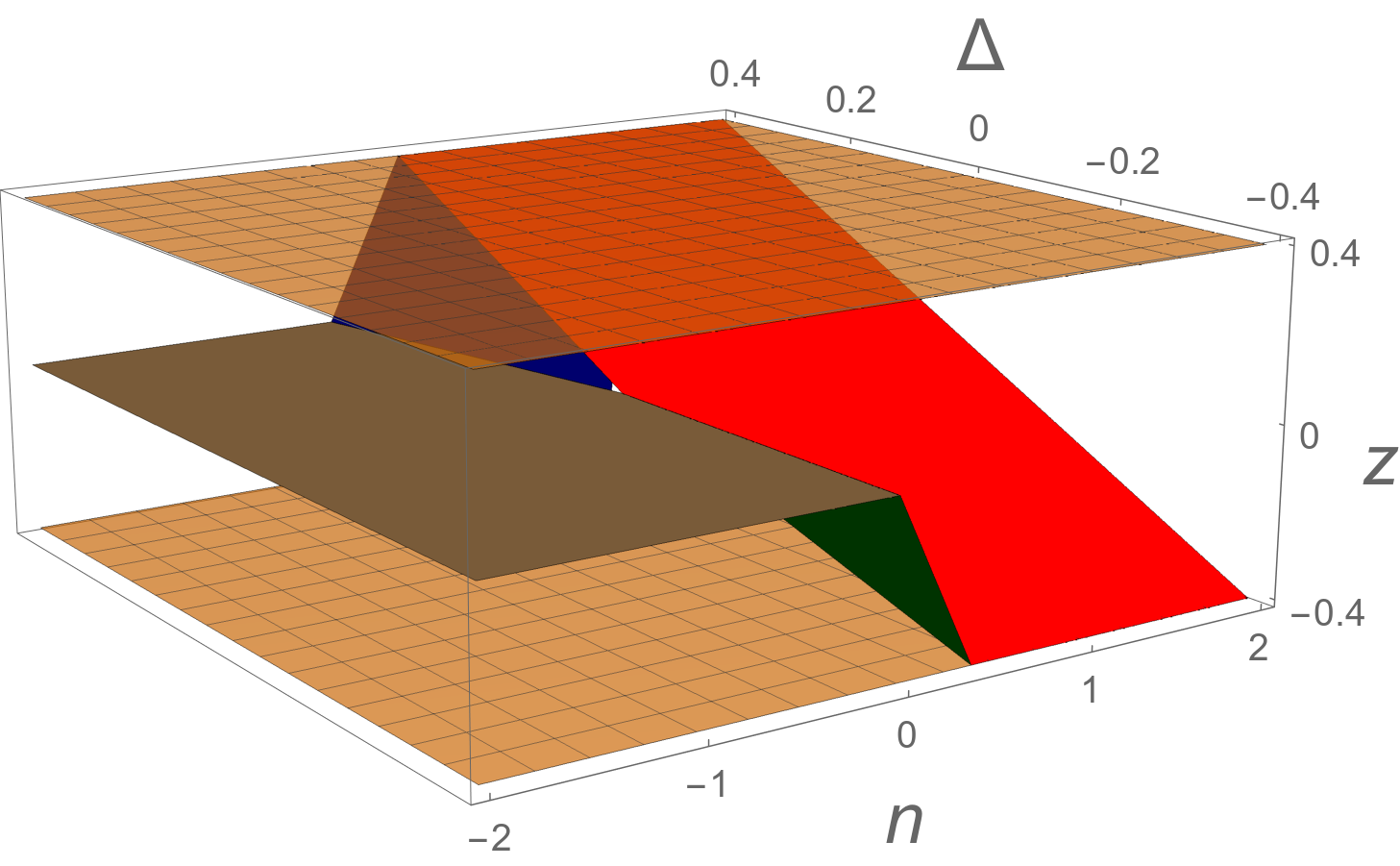}
\caption{Bounding planes computed by \ReluDiff{}~\cite{PaulsenWW20}.}
\label{fig:naiveboundingplanes}
\end{minipage}

\begin{minipage}[t]{0.45\linewidth}
\centering
%\captionsetup{width=0.98\textwidth}
\includegraphics[width=\linewidth]{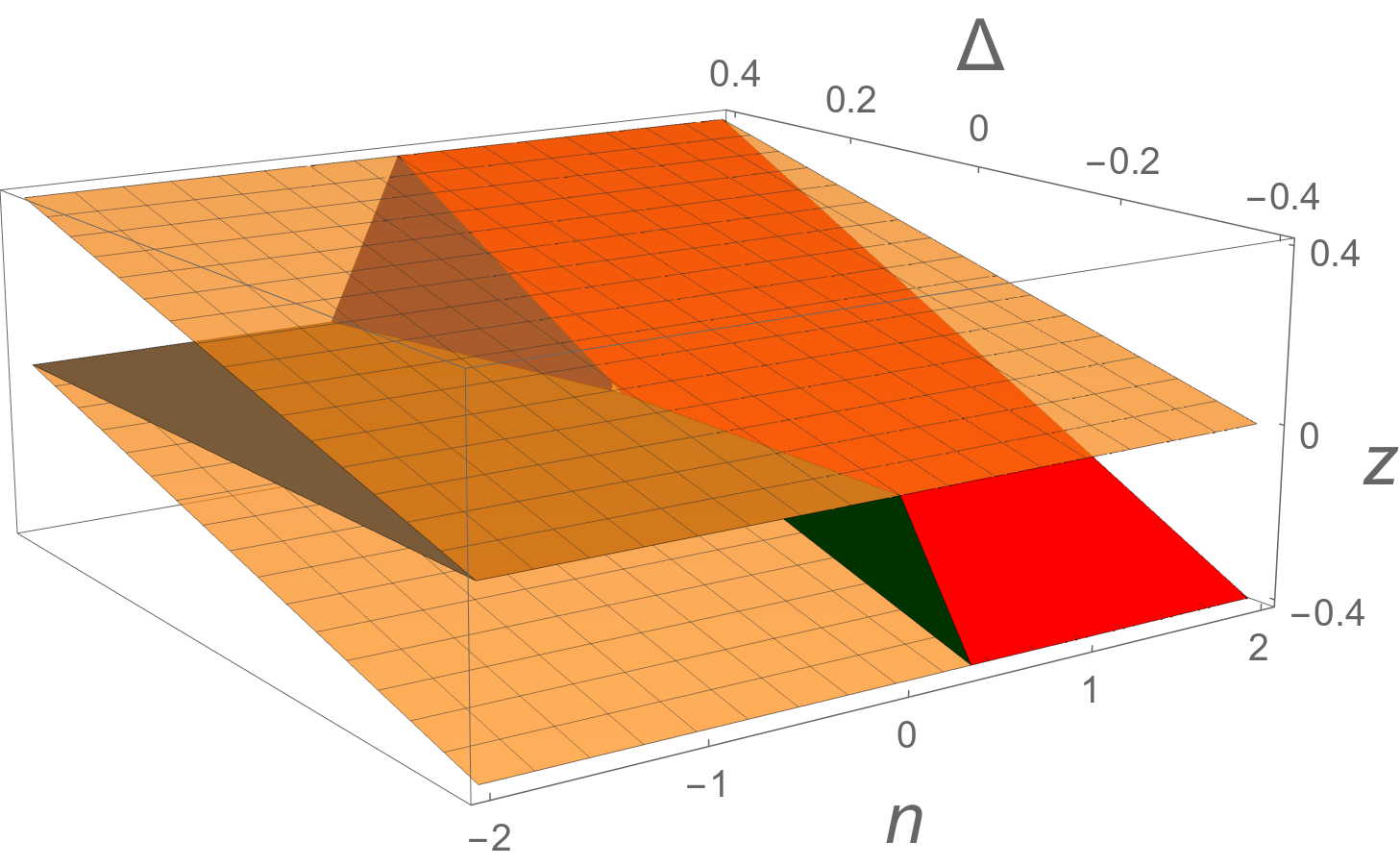}
\caption{Bounding planes computed by our new method.}
\label{fig:linearboundingplanes}
\end{minipage}%
\hspace{0.09\linewidth}
\begin{minipage}[t]{0.45\linewidth}
\centering
\scalebox{.65}{\tikzset{every picture/.style={line width=0.75pt}} %set default line width to 0.75pt        

\begin{tikzpicture}[x=0.75pt,y=0.75pt,yscale=-1,xscale=1]

%Straight Lines [id:da5621186100506654] 
\draw [color={rgb, 255:red, 199; green, 199; blue, 199 }  ,draw opacity=1 ]   (102,180) -- (338,180) ;
\draw [shift={(340,180)}, rotate = 180] [color={rgb, 255:red, 199; green, 199; blue, 199 }  ,draw opacity=1 ][line width=0.75]    (10.93,-3.29) .. controls (6.95,-1.4) and (3.31,-0.3) .. (0,0) .. controls (3.31,0.3) and (6.95,1.4) .. (10.93,3.29)   ;
\draw [shift={(100,180)}, rotate = 0] [color={rgb, 255:red, 199; green, 199; blue, 199 }  ,draw opacity=1 ][line width=0.75]    (10.93,-3.29) .. controls (6.95,-1.4) and (3.31,-0.3) .. (0,0) .. controls (3.31,0.3) and (6.95,1.4) .. (10.93,3.29)   ;
%Straight Lines [id:da9708804498380784] 
\draw [color={rgb, 255:red, 199; green, 199; blue, 199 }  ,draw opacity=1 ]   (220,62) -- (220,238) ;
\draw [shift={(220,240)}, rotate = 270] [color={rgb, 255:red, 199; green, 199; blue, 199 }  ,draw opacity=1 ][line width=0.75]    (10.93,-3.29) .. controls (6.95,-1.4) and (3.31,-0.3) .. (0,0) .. controls (3.31,0.3) and (6.95,1.4) .. (10.93,3.29)   ;
\draw [shift={(220,60)}, rotate = 90] [color={rgb, 255:red, 199; green, 199; blue, 199 }  ,draw opacity=1 ][line width=0.75]    (10.93,-3.29) .. controls (6.95,-1.4) and (3.31,-0.3) .. (0,0) .. controls (3.31,0.3) and (6.95,1.4) .. (10.93,3.29)   ;
%Straight Lines [id:da43894388755903013] 
\draw [color={rgb, 255:red, 74; green, 144; blue, 226 }  ,draw opacity=1 ][line width=1.5]    (220,180) -- (340,60) ;
%Straight Lines [id:da1261778605228432] 
\draw [color={rgb, 255:red, 74; green, 144; blue, 226 }  ,draw opacity=1 ][line width=1.5]    (220,180) -- (100,180) ;
%Straight Lines [id:da002818047409691604] 
\draw [color={rgb, 255:red, 0; green, 0; blue, 0 }  ,draw opacity=1 ] [dash pattern={on 4.5pt off 4.5pt}]  (320,80) -- (120,180) ;
%Straight Lines [id:da1715235527986837] 
\draw [color={rgb, 255:red, 0; green, 0; blue, 0 }  ,draw opacity=1 ] [dash pattern={on 4.5pt off 4.5pt}]  (320,130) -- (120,230) ;
%Straight Lines [id:da21391376020902975] 
\draw    (120,175.28) -- (120,180) ;
%Straight Lines [id:da9349071337756625] 
\draw    (120,180) -- (120,184.72) ;
%Straight Lines [id:da2699517570437805] 
\draw    (320,175.28) -- (320,180) ;
%Straight Lines [id:da4265291468096436] 
\draw    (320,180) -- (320,184.72) ;

% Text Node
\draw (117.5,196) node  [font=\Large]  {$\underline{LB}\left( n \right)$};
% Text Node
\draw (337,196) node  [font=\Large]  {$\overline{UB}\left( n \right)$};
% Text Node
\draw (166.5,140.5) node  [font=\Large,rotate=-333.19]  {$UB( ReLU(n) )$};
% Text Node
\draw (300,152.46) node  [font=\Large,rotate=-333.19]  {$LB( ReLU(n) )$};

\end{tikzpicture}}
\caption{Bounding planes computed by \Neurify{}~\cite{WangPWYJ18nips}.}
\label{fig:wangconvex}
\end{minipage}
\end{figure*}

The key contribution in \Name{}, our new method, is a \emph{symbolic}
and \emph{fine-grained} approximation technique that both reduces the
approximation error introduced when a neuron is in an unstable state,
and mitigates the explosion of such approximation error after it is
introduced.

\subsubsection{Convex Approximation for the Difference Interval}

Our first contribution is developing convex approximations to directly
bound the difference between two neurons after these ReLU activations.
Specifically, for a neuron $ n $ in $ f $ and corresponding neuron $
n' $ in $ f' $, we want to bound the value of $ \relu{n'} - \relu{n}
$. We illustrate the various choices using
Figures~\ref{fig:equation1}, \ref{fig:naiveboundingplanes},
and~\ref{fig:linearboundingplanes}.

The naive way to bound this difference is to first compute
approximations of $ y = \relu{n} $ and $ y' = \relu{n'} $ separately,
and then subtract them.  Since each of these functions has a single
variable, convex approximation is simple and is already used by
single-network verification tools~\cite{SinghGPV19,WangPWYJ18nips,WengZCSHDBD18}.
Figure~\ref{fig:wangconvex} shows the function $y=\relu{n}$ and its
bounding planes (shown as dashed-lines) in a two-dimensional space (details in
Section~\ref{sec:preliminary}).
However, as we have already mentioned, approximation errors would be
accumulated in the bounds of $\relu{n}$ and $\relu{n'}$ and then amplified
by the interval subtraction.  This is precisely why the naive approach
performs poorly.

The \ReluDiff{} method of Paulsen et al.~\cite{PaulsenWW20} improves
upon the new approximation by computing an interval bound on $ n' - n
$, denoted $ \Delta $, then rewriting $z = \relu{n'} - \relu{n} $ as $
z= \relu{n + \Delta} - \relu{n} $, and finally bounding this new
function instead.
Figure~\ref{fig:equation1} shows the shape of $ z = \relu{n
+ \Delta}- \relu{n} $ in a three-dimensional space. Note that it has
four piece-wise linear subregions, defined by values of the input
variables $n$ and $\Delta$.
While the bounds computed by \ReluDiff{}~\cite{PaulsenWW20}, shown as
the (horizontal) yellow planes in
Figure~\ref{fig:naiveboundingplanes}, are sound, in practice they tend
to be loose because the upper and lower bounds are both concrete
values.  Such eager concretization eliminates symbolic information
that $ \Delta $ contained before applying the ReLU activation.

In contrast, our method computes a convex approximation of $ z $,
shown by the (tilted) yellow planes in
Figure~\ref{fig:linearboundingplanes}.
Since these tilted bounding planes are in a three-dimensional space,
they are significantly more challenging to compute than the standard
two-dimensional convex approximations (shown in Figure~\ref{fig:wangconvex})
used by single network verification tools.
%convex approximation (shown in Figure~\ref{fig:wangconvex}) in a
%two-dimentionsl space, which was used by single-network verification
%tools.
%
Our approximations have the advantage of introducing significantly less
error than the horizontal planes used
in \ReluDiff{}~\cite{PaulsenWW20}, while maintaining some of the
symbolic information for $ \Delta $ before applying the ReLU
activation.

We will show through experimental evaluation
(Section~\ref{sec:experiment}) that our convex approximation can
drastically improve the accuracy of the difference bounds, and are
particularly effective when the input region being considered is
large.  Furthermore, the tilted planes shown in
Figure~\ref{fig:linearboundingplanes} are for the general case.  For
certain special cases, we obtain even tighter bounding planes (details
in Section~\ref{sec:approach}).
In the running example, using our new convex approximations would
improve the final bounds to $ [-1.97, 1.42] $.

\subsubsection{Symbolic Variables for Hidden Neurons}

Our second contribution is introducing symbolic variables to represent
the output values of some unstable neurons, with the goal of limiting
the propagation of approximation errors after they are introduced.
In the running example, since both $ \n{1}{1} $ and $ \np{1}{1} $ are in unstable states,
i.e., the input intervals of the ReLUs contain the value 0, we may introduce a
new
symbol $ x_3 = \relu{\np{1}{1}} - \relu{\n{1}{1}} $.  In all
subsequent layers, whenever the value of $\relu{\np{1}{1}}
- \relu{\n{1}{1}}$ is needed, we use the bounds $[x_3,x_3]$ instead of
the actual bounds.

The reason why using $x_3$ can lead to more accurate results is
because, even though our convex approximations reduce the error
introduced, there is inevitably some error that accumulates. Introducing $
x_3 $ allows this error to partially cancel in the subsequent
layers. In our running example, introducing the new symbolic variable
$x_3$ would be able to improve the final bounds to $ [-1.65, 1.18] $.

While creating $ x_3 $ improved the result
in this case, carelessly introducing new variables for all the unstable
neurons can actually reduce the overall benefit (see
Section~\ref{sec:approach}). In addition, the
computational cost of introducing new variables is not negligible.
%While substituting the values of unstable neurons with new symoblic
%variables may increases the accuracy of interval analysis, it may also
%add computational cost.
Therefore, in practice, we must introduce
these symbolic variables judiciously, to maximize the benefit.
Part of our contribution in \Name{} is in developing heuristics to
automatically determine when to create new symbolic variables (details
in Section~\ref{sec:approach}).

\section{Background}
\label{sec:preliminary}

In this section, we review the technical background and then introduce
notations that we use throughout the paper.

\subsection{Neural Networks}

% Define a neural network
% Limit ourselves to FFN ReLU networks
% Weight matrix notation
%	- W_k is the matrix for layer k
%	- W_k[i,j] is the weight of connection from n_k-1,i to n_k,j
We focus on feed-forward neural networks, which we define as a function $ f $ that takes
an $ n $-dimensional vector of real values $x \in \mathbb{X}$, where
$\mathbb{X} \subseteq \mathbb{R}^n $, and maps it to an $ m
$-dimensional vector $ y \in \mathbb{Y}$, where
$\mathbb{Y} \subseteq \mathbb{R}^m $.  We denote this function as $ f
: \mathbb{X} \to \mathbb{Y} $.  Typically, each dimension of $ y $
represents a score, such as a probability, that the input $ x $
belongs to class $ i $, where $ 1 \leq i \leq m $.

A network with $l$ layers has $ l $ weight matrices, each of which is
denoted $ \Wk{k}, $ for $ 1 \leq k \leq l $. For each weight matrix, we
have $ \Wk{k} \in \mathbb{R}^{l_{k-1} \times l_k} $ where $ l_{k-1} $
is the number of neurons in layer $ (k-1) $ and likewise for $ l_k $,
and $ l_0 = n $.
Each element in $ \Wk{k} $ represents the weight of an edge from a
neuron in layer $ (k-1) $ to one in layer $ k $. Let $ \n{k}{j} $
denote the $ j^{th} $ neuron of layer $ k $, and $ \n{k-1}{i} $ denote
the $ i^{th} $ neuron of layer $ (k-1) $.  We use $ \W{k}{i}{j} $ to
denote the edge weight from $ \n{k-1}{i} $ to $ \n{k}{j} $.
In our motivating example, we have $ W_1[1,1] = 1.9 $ and $ W_1[1,2] =
1.1 $.

Mathematically, the entire neural network can be represented by $ f(x)
= f_l(\, W_l \cdot \, f_{l-1}( \, W_{l-1} \, \cdot ...f_1(W_1 \cdot
x)...)) $, where $ f_k $ is the activation function of the $
k^{th} $ layer and $1\leq k\leq l$. We focus on neural networks with
ReLU activations because they are the most widely implemented in
practice, but our method can be extended to other activation
functions, such as $sigmoid$ and $tanh$, and other layer types, such
as convolutional and max-pooling. We leave this as future work.

\subsection{Symbolic Intervals}

% Introduce notation for intervals
To compute approximations of the output nodes that are sound for all
input values, we leverage interval
arithmetic~\cite{moore2009introduction}, which can be viewed as an
instance of the abstract interpretation framework~\cite{CousotC77}. It
is well-suited to the verification task because interval arithmetic is
soundly defined for basic operations of the network such as addition,
subtraction, and scaling.

Let $ I = [\LBEq{I}, \UBEq{I}] $ be an interval with lower bound
$ \LBEq{I} $ and upper bound $ \UBEq{I} $. Then, for intervals $ I_1,
I_2 $, we have addition and subtraction defined as $ I_1 + I_2 =
[\LBEq{I_1} + \LBEq{I_2}, \UBEq{I_1} + \UBEq{I_2}] $ and $ I_1 - I_2 =
[\LBEq{I_1} - \UBEq{I_2}, \UBEq{I_1} - \LBEq{I_2}] $, respectively.
For a constant $ c $, scaling is defined as $ c \times I_1 =
[c \times \LBEq{I_1}, c \times \UBEq{I_1} $ when $ c > 0 $, and $
c \times I_1 = [c \times \UBEq{I_1}, c \times \LBEq{I_1}] $ otherwise.

While interval arithmetic is a sound over-approximation, it is not
always accurate.  To illustrate, let $ f(x) = 3x - x $, and say we are
interested in bounding $ f(x) $ when $ x \in [-1, 1] $. One way to
bound $ f $ is by evaluating $ f(I) $ where $ I = [-1, 1] $. Doing so
yields $ 3 \times [-1, 1] - [-1, 1] = [-4, 4] $. Unfortunately, the most
accurate bounds are $ [-2, 2] $.

There are (at least) two ways we can improve the accuracy. First, we can
soundly refine the result by dividing the input intervals into disjoint
partitions, performing the analysis independently on each partition, and then
unioning the resulting output intervals together. Previous work has shown the
result will be \textit{at least} as precise~\cite{WangPWYJ18}, and often
better. For example, if we partition $ x \in [-1, 1] $ into $x\in[-1,0]$ and
$x\in[0,1]$, and perform the analysis for each partition, the
resulting bounds improve to $ [-3, 3] $.

% Introduce symblic interval concept and notation
Second, the dependence between the two intervals are
not leveraged when we subtract them, i.e., that they were both $ x $
terms and hence could partially cancel out.
To capture the dependence, we can use \textit{symbolic} lower and
upper bounds~\cite{WangPWYJ18}, which are expressions in
terms of the input variable, i.e., $ I = [x, x] $.  Evaluating $ f(I)
$ then yields the interval $ I_f = [2x, 2x] $, for $x\in[-1,1]$.  When using
symbolic bounds, eventually, we must concretize the lower and upper
bound equations. We denote concretization of $\LBEq{I_f}=2x$ and
$\UBEq{I_f}=2x$ as $ \LBL{I_f} = -2 $ and $ \UBU{I_f} = 2 $,
respectively.  Compared to the naive solution, $[-4,4]$, this is a
significant improvement.

When approximating the output of a given function $ f
: \mathbb{X} \to \mathbb{Y} $ over an input interval $
X \subseteq \mathbb{X} $, one may prove soundness by showing that the
evaluation of the lower and upper bounds on any input $ x \in X $ are
always greater than and less than, respectively, to the true value of
$ f(x) $.
Formally, for an interval $ I $, let $ \LBEq{I}(x) $ be the evaluation
of the lower bound equation on input $ x $, and similarly for
$ \UBEq{I}(x) $. Then, the approximation is considered sound if
$ \forall x \in X$, we have $\LBEq{I}(x) \leq f(x) \leq \UBEq{I}(x) $.

\subsection{Convex Approximations}
While symbolic intervals are exact for linear operations
(i.e. they do not introduce error), this is not the case
for non-linear operations, such as the ReLU activation.
%While using symbolic intervals in linear operations can avoid
%approximation errors, this is no longer possible for non-linear
%operations such as ReLU activation.
This is because, for efficiency
reasons, the symbolic lower and upper bounds must be kept linear.
Thus, developing linear approximations for non-linear activation
functions has become a signifciant area of research for single neural
network
verification~\cite{WangPWYJ18nips,SinghGPV19,WengZCSHDBD18,zhang2018efficient}. We
review the basics below, but caution that they are different from our
new convex approximations in \Name{}.

We denote the input to the ReLU of a neuron $ \nj $ as $ \IntIn{\nj} $
and the output as $ \IntOut{\nj} $. The approach used by existing
single-network verification tools is to apply an affine transformation
to the upper bound of $ \IntIn{\nj} $ such that
$ \UBEq{\IntIn{\nj}}(x) \geq 0 $, where $ x \in X $, and $ X $ is the
input region for the entire network.  For the lower bound, there exist
several possible transformations, including the one used
by \Neurify{}~\cite{WangPWYJ18nips}, shown in
Figure~\ref{fig:wangconvex}, where $n = \IntIn{\nj}$ and the dashed
lines are the upper and lower bounds.

We illustrate the upper bound transformation for $ \n{1}{1} $ of our
motivating example. After computing the upper bound of the ReLU
input $ \UBEq{\IntIn{\n{1}{1}}} = 1.9x_1 - 1.9x_2 $, where
$x_1\in[-2,2]$ and $x_2\in[-2,2]$, it computes the concrete lower and
upper bounds. We denote these as $ \UBL{\IntIn{\n{1}{1}}} = -7.6 $
and $ \UBU{\IntIn{\n{1}{1}}} = 7.6 $. We refer to them as $ l $ and $
u $, respectively, for short hand. Then, it computes the line that
passes through $ (u, u) $ and $ (0, l) $.
Letting $ y = \UBEq{\IntIn{\n{1}{1}}} $ be the upper bound equation of the ReLU
input, it computes the upper bound of the ReLU output as
$ \UBEq{\IntOut{\n{1}{1}}} $ $ = \frac{u}{u - l}(y - l) = $ $ 0.95x_1
-0.95x_2 +3.81 $.

When considering a single ReLU of a single network, convex
approximation is simple because there are only three states that the
neuron can be in, namely active, inactive, and unstable. Furthermore,
in only one of these states, convex approximation is needed.
In contrast, differential verification has to consider a pair of
neurons, which has up to nine states to consider between the two
ReLUs.  Furthermore, different states may result in different linear
approximations, and some states can even have multiple linear
approximations depending on the difference bound of $ \Delta = n' - n
$. As we will show in Section~\ref{sec:approach}, there are
significantly more considerations in our problem domain.

\ignore{
\subsection{Refinement}
% Interval analysis
Often times, the output interval will not be tight enough to verify the desired property. One way to improve the result is to divide the input interval into disjoint sub-intervals, perform the analysis on each region, and then union the output intervals together. The resulting output interval is guaranteed to be at least as tight the original output interval~\cite{moore2009introduction}. In fact, previous work showed that, for ReLU networks, the analysis result can be made arbitrarily accurate through input partitioning based refinement~\cite{WangPWYJ18}.

The key challenge in this type of refinement is deciding how to partition the input interval. Previous work~\cite{PaulsenWW20, WangPWYJ18, WangPWYJ18nips} has seen success by computing \textit{smear values}~\cite{kearfott1990algorithm, kearfott2013rigorous} for each input. Informally, the smear value is a measure of how much influence a single input neuron has on the output values.

For example, ....
}

\section{Our Approach}
\label{sec:approach}

We first present our baseline procedure for differential verification
of feed-forward neural networks (Section~\ref{sec:baseline}), and then
present our algorithms for computing convex approximations
(Section~\ref{sec:convex}) and introducing symbolic variables
(Section~\ref{sec:symbolic}).

\subsection{Differential Verification -- Baseline}
\label{sec:baseline}

We build off the work of Paulsen et al.~\cite{PaulsenWW20}, so in this section
we review the relevant pieces. We assume that the input
to \Name{} consists of two networks $ f $ and $ f' $, each with $ l $ layers
of the same size.  Let $ \npj $ in $ f' $ be the neuron paired with
$ \nj $ in $ f $. This implicitly creates a pairing of the edge
weights between the two networks. We first introduce additional notation.
\begin{itemize}
\item
We denote the difference between a pair of neurons as $ \ndj = \npj
- \nj $. For example, $ \nd{1}{1} = 0.1 $ under the input $ x_1 = 2, x_2 =
1 $ in our motivating example shown in Figure~\ref{fig:motex}.
\item
We denote the difference in a pair of edge weights as $ \Wd{k}{i}{j}
= \Wp{k}{i}{j} - \W{k}{i}{j} $. For example, $ \Wd{1}{1}{1} = 2.0 -
1.9 = 0.1 $.
\item
We extend the symbolic interval notation to these terms. That is,
$ \IntIn{\ndj} $ denotes the interval that bounds $ \npj - \nj $
before applying ReLU, and $ \IntOut{\ndj} $ denotes the interval after
applying ReLU.
\end{itemize}

Given that we have computed $ \IntOut{\n{k-1}{i}}$,
$\IntOut{\np{k-1}{i}}$, $\IntOut{\nd{k-1}{i}} $ for every neuron in
the layer $ k - 1 $, now, we compute a single $ \IntOut{\nd{k}{j}} $
in the subsequent layer $k$ in two steps (and then repeat for each $ 1 \leq
j \leq l_k $).

%\begin{enumerate}
%\item
First, we compute $ \IntIn{\nd{k}{j}} $ by propagating the output
intervals from the previous layer through the edges connecting to the
target neuron. This is defined as
\[
	\IntIn{\nd{k}{j}} = \sum_{i} \bigg( \IntOut{\nd{k-1}{i}} \times \Wp{k}{i}{j} + \IntOut{\n{k-1}{i}} \times \Wd{k-1}{i}{j} \bigg)
\]
We illustrate this computation on node $ \nd{1}{1} $ in our
example. First, we initialize $ \IntOut{\nd{0}{1}} = [0,0]$ ,
$\IntOut{\nd{0}{2}} = [0,0]$. Then we compute $ \IntIn{\nd{1}{1}} =
[0, 0] \times 2.0 +$$ [x_1, x_1] \times 0.1 + $$ [0, 0] \times -2.0 +
$ $ [x_2, x_2] \times -0.1 = $ $[0.1x_1 - 0.1x_2,$ $ 0.1x_1 - 0.1x_2]
$.

%\item
For the second step, we apply ReLU to $ \IntIn{\nd{k}{j}} $ to obtain
$ \IntOut{\nd{k}{j}} $. This is where we apply the new convex
approximations (Section~\ref{sec:convex}) to obtain tighter bounds.
Toward this end, we will focus on the following two equations:
\begin{align}
z_1 = \relu{{\nj} + {\ndj}} - \relu{{\nj}} \label{eq:1}\\
z_2 = \relu{{\npj}} - \relu{{\npj} - {\ndj}} \label{eq:2}
\end{align}
While Paulsen et al.~\cite{PaulsenWW20} also compute bounds of these
two equations, they use \emph{concretizations} instead of \emph{linear
approximations}, thus throwing away all the symbolic information.  For
the running example, their method would result in the bounds of
$ \IntOut{\nd{1}{1}} = [-.4, .4] $. In contrast, our method will be
able to maintain some or all of the symbolic information, thus
improving the accuracy.
%\end{enumerate}

\subsection{Two Useful Lemmas}

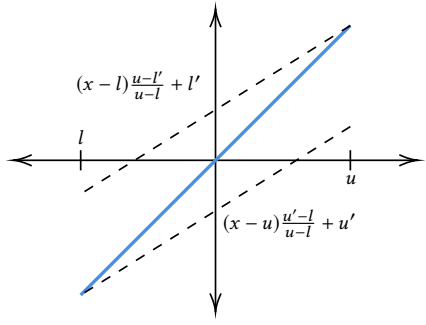
\begin{figure}[t]
	\scalebox{0.85}{\centering \tikzset{every picture/.style={line width=0.75pt}} %set default line width to 0.75pt        

\begin{tikzpicture}[x=0.75pt,y=0.75pt,yscale=-1,xscale=1]

%Straight Lines [id:da5621186100506654] 
\draw [color={rgb, 255:red, 0; green, 0; blue, 0 }  ,draw opacity=1 ][fill={rgb, 255:red, 0; green, 0; blue, 0 }  ,fill opacity=1 ]   (102,180) -- (338,180) ;
\draw [shift={(340,180)}, rotate = 180] [color={rgb, 255:red, 0; green, 0; blue, 0 }  ,draw opacity=1 ][line width=0.75]    (10.93,-3.29) .. controls (6.95,-1.4) and (3.31,-0.3) .. (0,0) .. controls (3.31,0.3) and (6.95,1.4) .. (10.93,3.29)   ;
\draw [shift={(100,180)}, rotate = 0] [color={rgb, 255:red, 0; green, 0; blue, 0 }  ,draw opacity=1 ][line width=0.75]    (10.93,-3.29) .. controls (6.95,-1.4) and (3.31,-0.3) .. (0,0) .. controls (3.31,0.3) and (6.95,1.4) .. (10.93,3.29)   ;
%Straight Lines [id:da9708804498380784] 
\draw [color={rgb, 255:red, 0; green, 0; blue, 0 }  ,draw opacity=1 ][fill={rgb, 255:red, 0; green, 0; blue, 0 }  ,fill opacity=1 ]   (220,92) -- (220,268) ;
\draw [shift={(220,270)}, rotate = 270] [color={rgb, 255:red, 0; green, 0; blue, 0 }  ,draw opacity=1 ][line width=0.75]    (10.93,-3.29) .. controls (6.95,-1.4) and (3.31,-0.3) .. (0,0) .. controls (3.31,0.3) and (6.95,1.4) .. (10.93,3.29)   ;
\draw [shift={(220,90)}, rotate = 90] [color={rgb, 255:red, 0; green, 0; blue, 0 }  ,draw opacity=1 ][line width=0.75]    (10.93,-3.29) .. controls (6.95,-1.4) and (3.31,-0.3) .. (0,0) .. controls (3.31,0.3) and (6.95,1.4) .. (10.93,3.29)   ;
%Straight Lines [id:da43894388755903013] 
\draw [color={rgb, 255:red, 74; green, 144; blue, 226 }  ,draw opacity=1 ][line width=1.5]    (140,260) -- (300,100) ;
%Straight Lines [id:da002818047409691604] 
\draw [color={rgb, 255:red, 0; green, 0; blue, 0 }  ,draw opacity=1 ] [dash pattern={on 4.5pt off 4.5pt}]  (300,100) -- (140,200) ;
%Straight Lines [id:da1715235527986837] 
\draw [color={rgb, 255:red, 0; green, 0; blue, 0 }  ,draw opacity=1 ] [dash pattern={on 4.5pt off 4.5pt}]  (300,160) -- (140,260) ;
%Straight Lines [id:da5537839721735281] 
\draw    (140,174) -- (140,186) ;
%Straight Lines [id:da6486767585115278] 
\draw    (300,174) -- (300,186) ;

% Text Node
\draw (136,125.4) node [anchor=north west][inner sep=0.75pt]    {$( x-l)\frac{u-l'}{u-l} +l'$};
% Text Node
\draw (223,207.4) node [anchor=north west][inner sep=0.75pt]    {$( x-u)\frac{u'-l}{u-l} +u'$};
% Text Node
\draw (137,160.4) node [anchor=north west][inner sep=0.75pt]    {$l$};
% Text Node
\draw (296,187.4) node [anchor=north west][inner sep=0.75pt]    {$u$};

\end{tikzpicture}}
	\caption{Illustration of Lemmas~\ref{lemma:1} and~\ref{lemma:2}.\label{fig:lemma_fig}}
\end{figure}

Before presenting our new linear approximations, we introduce two
useful lemmas, which will simplify our presentation as well as our
soundness proofs.

\begin{lemma}\label{lemma:1}
	Let $ x $ be a variable such that $ l \leq x \leq u $ for constants $ l \leq 0 $ and $ 0 \leq u $. For a constant $ l' $ such that $ l \leq l' \leq 0 $, we have $ x \leq (x - l) * \frac{u - l'}{u - l} + l' \geq l' $.
\end{lemma}
\begin{lemma}\label{lemma:2}
	Let $ x $ be a variable such that $ l \leq x \leq u $ for constants $ l \leq 0 $ and $ 0 \leq u $. For a constant $ u' $ such that $ 0 \leq u' \leq u $, we have $ u' \geq (x - u) * \frac{u' - l}{u - l} + u' \leq x $.
\end{lemma}

We illustrate these lemmas in Figure~\ref{fig:lemma_fig}. The solid
blue line shows the equation $ y = x $ for the input interval $ l \leq
x \leq u $. The upper dashed line illustrates the transformation of
Lemma~\ref{lemma:1}, and the lower dashed line illustrates
Lemma~\ref{lemma:2}. Specifically, Lemma~\ref{lemma:1} shows a
transformation applied to $ x $ whose result is always greater than
both $ l' $ and $ x $. Similarly, Lemma~\ref{lemma:2} shows a
transformation applied to $ x $ whose result is always less than both
$ u' $ and $ x $. These lemmas will be useful in bounding Equations~\ref{eq:1}
and~\ref{eq:2}.

\subsection{New Convex Approximations for $ \IntOut{\ndj} $}
\label{sec:convex}

Now, we are ready to present our new approximations, which are
linear symbolic expressions derived from Equations~\ref{eq:1}
and~\ref{eq:2}.

We first assume that $ \nj $ and $ \npj $ could both be unstable,
i.e., they could take values both greater than and less than 0. This
yields bounds for the general case in that they are sound in all
states of $ \nj $ and $ \npj $ (Sections~\ref{sec:upper}
and \ref{sec:lower}).
Then, we consider special cases of $ \nj $ and $ \npj $, in
which even tighter upper and lower bounds are derived
(Section~\ref{sec:tighter}).

\newcommand{\nn}[0]{n}
\newcommand{\nnp}[0]{n'}
\newcommand{\nnd}[0]{\Delta}

To simplify notation, we let $ \nn, \nnp,$ and $ \nnd $ stand in for $ \nj,
\npj, $ and $ \ndj $ in the remainder of this section.

\subsubsection{Upper Bound for the General Case}
\label{sec:upper}

Let $ l = \UBL{\IntIn{\nnd}}$ and $ u = \UBU{\IntIn{\nnd}} $. The
upper bound approximation is:
\[
\UBEq{\IntOut{\nnd}} = \begin{cases}
\UBEq{\IntIn{\nnd}} & \UBL{\IntIn{\nnd}} \geq 0\\
0 & \UBU{\IntIn{\nnd}} \leq 0 \\
(\UBEq{\IntIn{\nnd}} - l)*\frac{u}{u - l} & otherwise
\end{cases}
\]
That is, when the input's (delta) upper bound is greater than 0 for all $
x \in X $, we can use the input's upper bound unchanged.  When the
upper bound is always less than 0, the new output's upper
bound is then 0. Otherwise, we apply a linear transformation to the
upper bound, which results in the upper plane illustrated in
Figure~\ref{fig:linearboundingplanes}. We prove all three cases sound.

\begin{proof}

We consider each case above separately. In the following, we use
Equation~\ref{eq:1} to derive the bounds, but we note a symmetric proof using
Equation~\ref{eq:2} exists and produces the same bounds.

\paragraph{Case 1: $ \UBL{\IntIn{\nnd}} \geq 0 $.}
We first show that, according to Equation~\ref{eq:1}, when $ 0 \leq
\nnd $ we have $ z_1 \leq \nnd $. This then implies that, if
$ \UBL{\IntIn{\nnd}} \geq 0 $, then $ z_1 \leq \UBEq{\IntIn{\nnd}}(x)
$ for all $ x \in X $, and hence it is a valid upper bound for the output
interval.

Assume $ 0 \leq \nnd $. We consider two cases of $ \nn $. First, consider
$ 0 \leq \nn $. Observe $ 0 \leq \nn \wedge 0 \leq \nnd $ $ \implies $
$ 0 \leq \nn + \nnd $. Thus, the ReLU's of Equation~\ref{eq:1}
simplify to $ z_1 = \nn + \nnd - \nn = \nnd \implies z_1 \leq \nnd
$. When $ \nn < 0 $, Equation~\ref{eq:1} simplifies to $ z_1
= \relu{\nn + \nnd} $. Since $ \nn < 0 $, we have $ \nn
+ \nnd \leq \nnd \wedge 0 \leq \Delta $ $ \implies $ $ \relu{\nn + \nnd}
\leq \nnd$. Thus,
$ z_1 = \relu{\nn + \nnd} \leq \nnd $, so the approximation is
sound.

\paragraph{Case 2: $ \UBU{\IntIn{\nnd}} \leq 0 $.}
This case was previously proven~\cite{PaulsenWW20}, but we restate it
here.  $ \UBU{\IntIn{\nnd}} \leq 0 $ $ \iff $ $ \nnp \leq \nn $ $ \implies $
$\relu{\nnp} \leq \relu{\nn} $
$ \iff $ $ \relu{\nnp} - \relu{\nn} \leq 0 $.

\paragraph{Case 3.}
By case 1, any $ \UBEq{\IntOut{\nnd}} $ that satisfies
$ \UBEq{\IntOut{\nnd}}(x) $ $ \geq 0 $ and $ \UBEq{\IntOut{\nnd}}(x) \geq
$ $ \UBEq{\IntIn{\nnd}}(x) $ for all $ x \in X $ is sound. Both
inequalities hold by Lemma~\ref{lemma:1}, with $ x
= \UBEq{\IntIn{\nnd}} $, $l=\UBL{\IntIn{\nnd}}$,
$u=\UBU{\IntIn{\nnd}}$ and $ l' = 0 $.

\end{proof}

We illustrate the upper bound computation on node $ \n{1}{1} $ of our
motivating example. Recall that $ \UBEq{\IntIn{\n{1}{1}}} = 0.1x_1 -
0.1x_2 $. Since $ \UBL{\IntIn{\n{1}{1}}} = -0.4 $ and
$ \UBU{\IntIn{\n{1}{1}}} = 0.4 $, we are in the third case of our
linear approximation above. Thus, we have $ \UBEq{\IntIn{\n{1}{1}}}
= $$ ( 0.1x_1 - 0.1x_2 - (-0.4))*\frac{0.4}{0.4 - (-0.4)} = $$ 0.5x_1
- 0.5x_2 + 0.2 $. This is the upper bounding plane illustrated in
Figure~\ref{fig:linearboundingplanes}. The volume under this plane is 50\% less
than the upper bounding plane of \ReluDiff{} shown in
Figure~\ref{fig:naiveboundingplanes}.

\subsubsection{Lower Bound for the General Case}
\label{sec:lower}

Let $ l = \LBL{\IntIn{\nnd}} $ and $ u = $ $ \LBU{\IntIn{\nnd}} $, the
lower bound approximation  is:
\[
\LBEq{\IntOut{\nnd}} = \begin{cases}
\LBEq{\IntIn{\nnd}} 	& 		\LBU{\IntIn{\nnd}} \leq 0\\
0 				& 		\LBL{\IntIn{\nnd}} \geq 0 \\
(\LBEq{\IntIn{\nnd}} - u)*\frac{-l}{u - l} & otherwise
\end{cases}
\]
That is, when the input lower bound is always less than 0, we can
leave it unchanged. When it is always greater than 0, the new lower
bound is then 0.  Otherwise, we apply a transformation to the lower
bound, which results in the lower plane illustrated in
Figure~\ref{fig:linearboundingplanes}. We prove all three cases sound.

\begin{proof}
We consider each case above separately. In the following, we use
Equation~\ref{eq:1} to derive the bounds, but we note a symmetric proof using
Equation~\ref{eq:2} exists and produces the same bounds.

\paragraph{Case 1: $ \LBU{\IntIn{\nnd}} \leq 0 $.}
We first show that according to Equation~\ref{eq:1}, when $ \nnd \leq 0 $ we
have $ \nnd \leq z_1 $. This then implies that, if $ \LBU{\IntIn{\nnd}} \leq 0
$, we have $ \LBEq{\IntIn{\nnd}}(x) \leq z_1 $ for all $ x \in X $, and hence
it is a valid lower bound for the output interval.

Assume $ \nnd \leq 0 $. We consider two cases of $ \nn + \nnd
$. First, let $ 0 \leq $ $ \nn + \nnd $. Observe $ 0 \leq $ $ \nn
+ \nnd \wedge \nnd \leq 0 $ $ \implies $ $ 0 \leq \nn $, so we can
simplify Equation~\ref{eq:1} to $ z_1 = \nn + \nnd - \nn
= \nnd \implies \nnd \leq z_1 $. Second, let $ \nn + \nnd <
0 \iff \nnd < -\nn $. Then, Equation~\ref{eq:1} simplifies to $ z_1 =
- \relu{\nn} = -max(0,\nn) = min(0, -\nn) $. Now observe $ \nnd <
-\nn \wedge \nnd < 0 \implies \nnd < min(0, -\nn) = z_1 $.

\paragraph{Case 2: $ \LBL{\IntIn{\nnd}} \geq 0 $.}
This case was previously proven sound \cite{PaulsenWW20}, but we
restate it here. $ \LBL{\IntIn{\nnd}} \geq 0 $ $ \iff $ $ \nnp \geq \nn $ $
\implies $ $ \relu{\nnp} \geq \relu{\nn} $ $ \iff $ $ \relu{\nnp} - \relu{\nn}
\geq 0 $.

\paragraph{Case 3.}
By case 1, any $ \LBEq{\IntOut{\nnd}} $ that satisfies
$ \LBEq{\IntOut{\nnd}}(x) \leq 0 $ and $ \LBEq{\IntOut{\nnd}}(x) $
$ \leq $ $ \LBEq{\IntIn{\nnd}}(x) $ for all $ x \in X $ will be valid. Both
inequalities hold by Lemma~\ref{lemma:2}, with $ x = \LBEq{\IntIn{\nnd}}, u'
= 0, l = \LBL{\IntIn{\nnd}}, $ and $ u = \LBU{\IntIn{\nnd}} $.

\end{proof}

We illustrate the lower bound computation on node $ \n{1}{1} $ of our
motivating example. Recall that $ \LBEq{\IntIn{\n{1}{1}}} = 0.1x_1 -
0.1x_2 $. Since $ \LBL{\IntIn{\n{1}{1}}} = -0.4 $ and
$ \LBU{\IntIn{\n{1}{1}}} = 0.4 $, we are in the third case of our
linear approximation. Thus, we have $ \LBEq{\IntOut{\n{1}{1}}} = $ $
(0.1x_1 - 0.1x_2 - (-0.4))*\frac{-(-0.4)}{0.4 - (-0.4)} $$ = 0.05x_1 -
0.05x_2 - 0.2$. This is the lower bounding plane illustrated in
Figure~\ref{fig:linearboundingplanes}. The volume above this plane is 50\% less
than the lower bounding plane of \ReluDiff{} shown in
Figure~\ref{fig:naiveboundingplanes}.

\subsubsection{Tighter Bounds for Special Cases}
\label{sec:tighter}

While the bounds presented so far apply in all states of $ \nn $ and
$ \nnp $, under certain conditions, we are able to tighten these
bounds even further.
Toward this end, we restate the following two lemmas proved by Paulsen
et al.~\cite{PaulsenWW20}, which will come in handy.  They are related
to properties of Equations~\ref{eq:1} and \ref{eq:2}, respectively.
\begin{lemma} \label{lemma:prev1}
	$ \relu{{\nn} + {\nnd}} - \nn \equiv max(-\nn, \nnd)$
\end{lemma}
\begin{lemma} \label{lemma:prev2}
	$ \nnp - \relu{{\nnp} - {\nnd}} \equiv min(\nnp, \nnd)$
\end{lemma}
These lemmas provide bounds when $ \nn $ and $ \nnp $ are proved to be
linear based on the absolute bounds that we compute.

\paragraph{Tighter Upper Bound When $ \nnp $ Is Linear.}

In this case, we have $ \UBEq{\IntOut{\nnd}} $ $ = $
$ \UBEq{\IntIn{\nnd}}$, which is an improvement for the second or
third case of our general upper bound.

\begin{proof}
By our case assumption, Equation~\ref{eq:2} simplifies to the one in
Lemma~\ref{lemma:prev2}. Thus, $ z_2 =$ $ min(\nnp, \nnd) $ $\implies
z_2 \leq \nnd $.
\end{proof}

\begin{figure}
	\includegraphics[width=\linewidth]{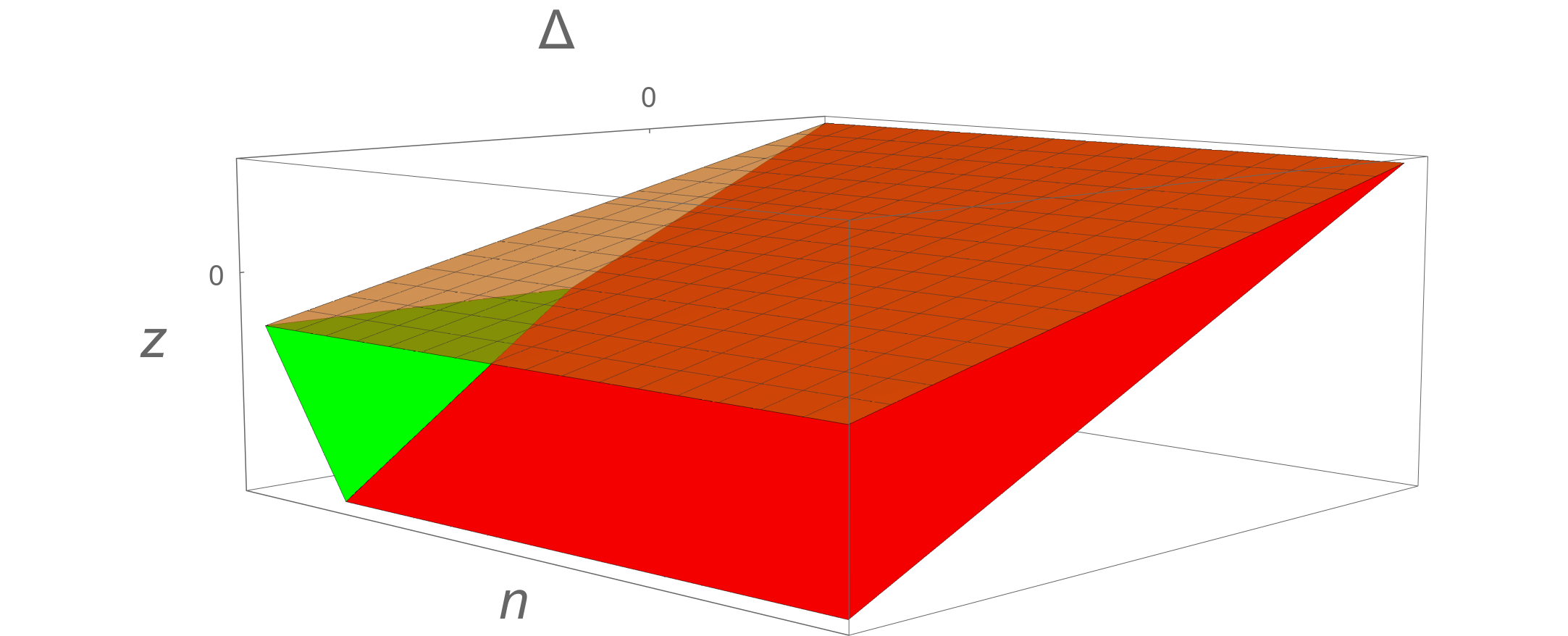}
	\caption{Tighter upper bounding plane.\label{fig:specialcase1}}
\end{figure}

\begin{figure}
	\includegraphics[width=\linewidth]{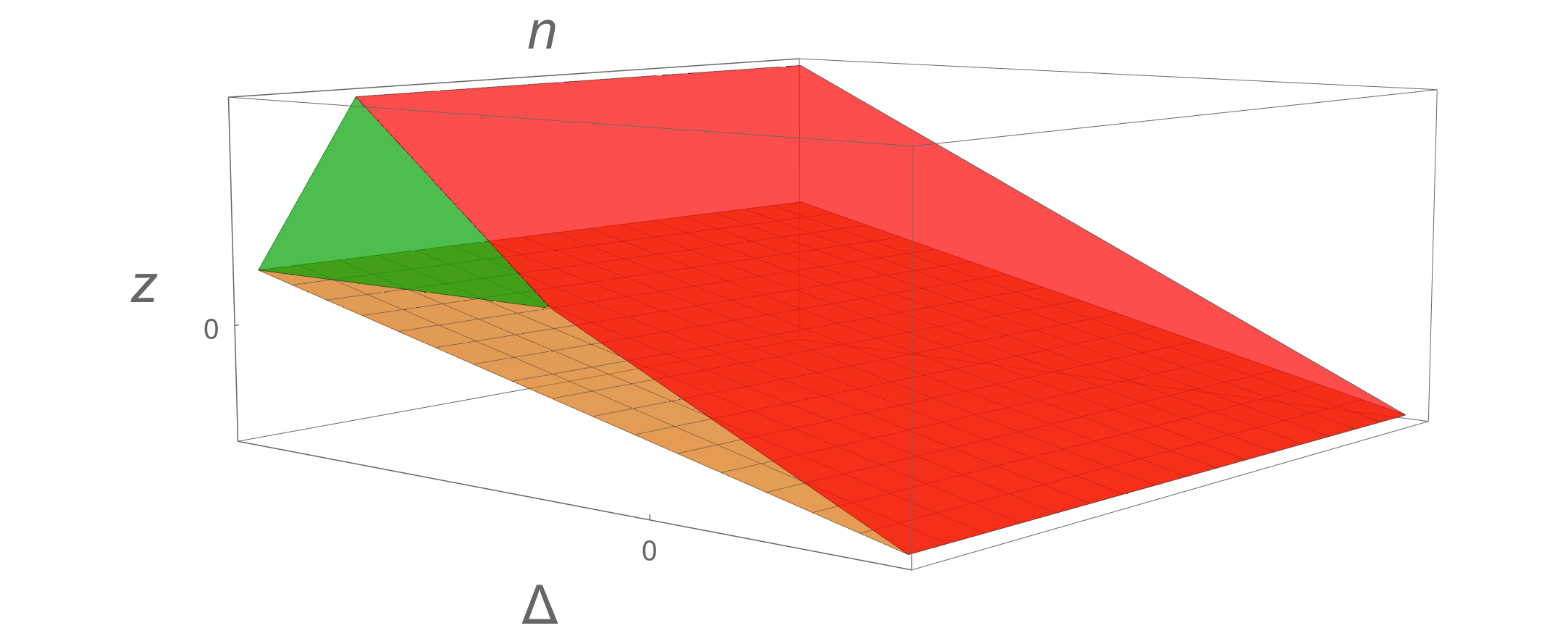}
	\caption{Tighter lower bounding plane.\label{fig:specialcase2}}
\end{figure}

\paragraph{Tighter Upper Bound When $ \nn $ Is Linear, $ \UBL{\IntIn{\nnd}}
\leq $  $-\LBL{\IntIn{\nn}} $ $ \leq \UBU{\IntIn{\nnd}} $.}
We illustrate the $ z_1 $ plane under these constraints in
Figure~\ref{fig:specialcase1}.
Let $ l = \UBL{\IntIn{\nnd}} $, and let $ u = \UBU{\IntIn{\nnd}} $, and $ l' =
-\LBL{\IntIn{\nn}} $, we use Lemma~\ref{lemma:1} to derive
$ \UBEq{\IntOut{\nnd}} =$ $ (\UBEq{\IntIn{\nnd}} - l)*\frac{u - l'}{u
- l} + l' $. This results in the upper plane of Figure~\ref{fig:specialcase1}.
This improves over the third case in our general upper
bound because it allows the lower bound of $ \UBEq{\IntOut{\nnd}} $ to
be less than 0.

\begin{proof}
By our case assumption, Equation~\ref{eq:1} simplifies to the one in
Lemma~\ref{lemma:prev1}. By Lemma~\ref{lemma:1}, we have for all $ x \in X $,
$ \UBEq{\IntOut{\nnd}}(x) \geq -\LBL{\IntIn{\nn}} $ and
$ \UBEq{\IntOut{\nnd}}(x) \geq \UBEq{\IntIn{\nnd}}(x) $. These two inequalities
imply $ \UBEq{\IntOut{\nnd}} \geq max(-\nn, \nnd) $.
\end{proof}

\paragraph{Tighter Lower Bound When $ \nn $ Is Linear.}

Here, we can use the approximation $ \LBEq{\IntOut{\nnd}}
= \LBEq{\IntIn{\nnd}} $. This improves over the second and third cases
of our general lower bound.

\begin{proof}
By our case assumption, Equation~\ref{eq:1} simplifies to the one in
Lemma~\ref{lemma:prev1}. Thus, $ z_1 = max(-\nn, \nnd) $ $ \implies $
$ z_1 \geq \nnd $.
\end{proof}

\paragraph{Tighter Lower Bound when $ \nnp $ is Linear, $ \LBL{\IntIn{\nnd}}
$ $ \leq \LBL{\IntIn{\nnp}} $ $ \leq \LBU{\IntIn{\nnd}}$.}
We illustrate the $ z_2 $ plane under these constraints in
Figure~\ref{fig:specialcase2}.
Here, letting $ l = \LBL{\IntIn{\nnd}} $, $ u = \LBU{\IntIn{\nnd}} $,
and $ u' = \LBL{\IntIn{\nnp}} $, we can use Lemma~\ref{lemma:2} to
derive the approximation $ \LBEq{\IntOut{\nnd}} $ $ =
(\LBEq{\IntIn{\nnd}} - u)*\frac{u - l'}{u - l} + u' $. This results in the
lower plane of Figure~\ref{fig:specialcase2}. This improves
over the third case, since it allows the upper bound to be greater
than 0.

\begin{proof}
By our case assumption, Equation~\ref{eq:2} simplifies to the one shown in
Lemma~\ref{lemma:prev2}. By Lemma~\ref{lemma:2}, we have for all $ x \in X $,
$ \LBEq{\IntOut{\nnd}}(x) \leq \LBEq{\IntIn{\nnd}}(x) $ and
$ \LBEq{\IntOut{\nnd}}(x) \leq \LBL{\IntIn{\nnp}} $. These two inequalities
imply
$ \LBEq{\IntOut{\nnd}}(x)$ $ \leq min(\nnp, \nnd) $.
\end{proof}

\subsection{Intermediate Symbolic Variables for $\IntOut{\nnd}$}
\label{sec:symbolic}

While convex approximations reduce the error introduced by ReLU, even
small errors tend to be amplified significantly after a few layers.
To combat the error explosion, we introduce new symbolic terms to represent the
output values of unstable neurons, which allow their accumulated
errors to cancel out.

We illustrate the impact of symbolic variables on $ \n{1}{1} $ of our
motivating example. Recall we have $ \IntOut{\nd{1}{1}} =$$ [0.05x_1 -
0.05x_2 - 0.2,$ $ 0.05x_1 - 0.05x_2 + 0.2] $. After applying the
convex approximation, we introduce a new variable $ x_3$
such that $ x_3 = $ $ [0.05x_1 - 0.05x_2 - 0.2, $ $
0.05x_1 - 0.05x_2 + 0.2 ]$.
Then we set $ \IntOut{\nd{1}{1}} = [x_3, x_3] $, and
propagate this interval as before. After propagating through
$ \n{2}{1} $ and $ \n{2}{2} $ and combining them at $ \n{3}{1} $, the
$ x_3 $ terms partially cancel out, resulting in the tighter final
output interval $ [-1.65, 1.18] $.

In principle, symbolic variables may be introduced at any unstable
neurons that introduce approximation errors,
however there are efficiency vs. accuracy tradeoffs when
introducing these symbolic variables.  One consideration
is how to deal with intermediate variables referencing other
intermediate variables. For example, if we decide to introduce a
variable $ x_4 $ for $ \n{2}{1} $, then $ x_4 $ will have an $ x_3 $
term in its equation. Then, when we are evaluating a symbolic bound that
contains
an $ x_4 $ term, which will be the case for $ \n{3}{1} $, we will have to
recursively substitute the bounds of the
previous intermediate variables, such as $ x_3 $. This becomes expensive,
especially when it is used together with our bisection-based
refinement~\cite{WangPWYJ18,PaulsenWW20}.
Thus, in practice, we first remove any back-references to intermediate
variables by substituting in their lower bounds and upper bounds into the
new intermediate variable's lower and upper bounds, respectively.

Given that we do not allow back-references, there are two
additional considerations.
First, we must consider that introducing a new intermediate
variable wipes out all the other intermediate variables.  For
example, introducing a new variable at $ \n{2}{1} $ wipes out references
to $ x_3 $, thus preventing any $ x_3 $ terms from canceling at $ \n{3}{1} $.
Second, the runtime cost of introducing symbolic variables is not
negligible. The bulk of computation time in \Name{} is spent
multiplying the network's weight matrices by the neuron's symbolic
bound equations, which is implemented using matrix multiplication. Since adding
variables increases the matrix size, this increases the matrix
multiplication cost.

Based on these considerations, we have developed heuristics for adding
new variables judiciously.
First, since the errors introduced by unstable neurons in
the \emph{earliest} layers are the most prone to explode, and hence
benefit the most when we create variables for them, we
rank them higher when choosing where to add symbolic variables.
Second, we bound the total number of symbolic variables that may be
added, since our experience shows that introducing symbolic variables
for the earliest $ N $ unstable neurons gives drastic improvements in
both run time and accuracy.
In practice, $N$ is set to a number proportional
to the weighted sum of unstable neurons in all layers. Formally,
$N=\Sigma_{k=1}^L \gamma^k \times N_k$, where $N_k$ is the number of
unstable neurons in layer $k$ and $\gamma^k = \frac{1}{k}$ is the discount
factor.
%
%\cwnote{Seriously, we have to fix $N$ in our tool, one way or
%another.  The reason is becuase changing $N$ for each benchmark
%arbitrarily sounds really bad!}

\section{Experiments}
\label{sec:experiment}

% Experiments we want to cover
% - ACAS
% 	- eps = 0.05, 0.01
%	- One big table or two small tables
% - MNIST
% 	- 4x1024
%		- increasing perturbation
%		- increasing number of pixels
%		- accuracy on perturb = 8
%	- 6x500
%		- increasing perturbation
%	- Have one line graph for increasing perturbs on both arches
%	- Also do an accuracy comparison for both arches

We have implemented \Name{} and compared it
with \ReluDiff{}~\cite{PaulsenWW20}, the state-of-the-art tool for
differential verification of neural networks.
\Name{} builds upon the codebase of \ReluDiff{}~\cite{reludiffrepo},
which was also used by single-network verification tools such
as \ReluVal{}~\cite{WangPWYJ18} and \Neurify{}~\cite{WangPWYJ18nips}.
All use OpenBLAS~\cite{ZhangWZ12} to optimize the symbolic
interval arithmetic (namely in applying the weight matrices to the
symbolic intervals).
We note that \Name{} uses the algorithm from \Neurify{} to compute
$ \IntOut{\nj} $ and $ \IntOut{\npj} $, whereas \ReluDiff{} uses
the algorithm of \ReluVal{}. Since \Neurify{} is known to compute
tighter bounds than \ReluVal{}~\cite{WangPWYJ18nips},
we compare to both \ReluDiff{}, and an upgraded version of \ReluDiff{}
which uses the bounds from \Neurify{} to ensure
that any performance gain is due to our optimizations and not due to
using \Neurify{}'s bounds. We use the name \ReluDiffP{} to refer
to \ReluDiff{} upgraded with \Neurify{}'s bounds.
%We also
%improved \ReluDiff{} to use a more accurate algorithm for computing
%the absolute values: the original \ReluDiff{} uses the algorithm
%from \ReluVal{}~\cite{WangPWYJ18} to compute $ \IntOut{\nj} $ and
%$ \IntOut{\npj} $, whereas the improved \ReluDiff{} uses the more
%accurate algorithm from \Neurify{}~\cite{WangPWYJ18nips}.
%
%Since our \Name{} also uses the more accurate algorithm
%from \Neurify{}, upgrading \ReluDiff{} to \ReluDiffP{} (the new
%version) allows us to have a more fair experimental comparison.

%To sum up, both \ReluDiff{}/\ReluDiffP{} and \Name{} take two
%structurally similiar networks $ f $ and $ f' $, an input region $ X
%$, and a small $ \epsilon $ value. They then attempt to prove that
%$ \forall x \in X$. $|f'(x) - f(x)| < \epsilon $ for each of the
%network's output nodes.

\subsection{Benchmarks}

Our benchmarks consist of the 49 feed-forward neural networks used by
Paulsen et al.~\cite{PaulsenWW20}, taken from three applications:
aircraft collision avoidance, image classification, and human activity
recognition. We briefly describe them here. As in Paulsen et al.~\cite{PaulsenWW20}, the second
network $ f' $ is generated by truncating the edge weights of $f$ from
32 bit to 16 bit floats.

\paragraph{ACAS Xu~\cite{JulianKO18}}
ACAS (aircraft collision avoidance system) Xu is a set of
forty-five neural networks, each with five inputs, six hidden layers of 50 units each, and five outputs,
designed to advise a pilot (the ownship) how to steer an aircraft in the presence of an intruder aircraft.
The inputs describe the position and speed of the intruder relative to the ownship, and the
outputs represent scores for different actions that the ownship should take.
The scores range from $ [-0.5, 0.5] $. We use the input regions defined by the properties of previous
work~\cite{KatzBDJK17, WangPWYJ18}.

\paragraph{MNIST~\cite{LecunBBH98}}
MNIST is a standard image classification task, where the goal is to
correctly classify $ 28 \times 28 $ pixel greyscale images of handwritten
digits. Neural networks trained for this task take 784 inputs (one for
each pixel) each in the range $ [0,255] $, and compute ten outputs -- one
score for each of the ten possible digits. We use three networks of size
3x100 (three hidden layers of 100 neurons each), 2x512, and 4x1024 taken from Weng et al.~\cite{WengZCSHDBD18} and
Wang et al.\cite{WangPWYJ18nips}. All achieve at least 95\% accuracy on
holdout test data.

\paragraph{Human Activity Recognition (HAR)~\cite{AnguitaHAR}}
The goal for this task is to classify
the current activity of human (e.g. walking, sitting, laying down) based
on statistics from a smartphone's gyroscopic sensors. Networks trained
on this task take 561 statistics computed from the sensors and output six
scores for six different activities. We use a 1x500 network.

\subsection{Experimental Setup}

Our experiments aim to answer the following research questions:

\begin{enumerate}
	\item Is \Name{} significantly faster than state-of-the-art?
	\item Is \Name{}'s forward pass significantly more accurate?
	\item Can \Name{} handle significantly larger input regions?
	\item How much does each technique contribute to the overall improvement?
\end{enumerate}

To answer these questions, we run both \Name{}
and \ReluDiff{}/\ReluDiffP{} on all benchmarks and compare their
results.  Both \Name{} and \ReluDiff{}/\ReluDiffP{} can be
parallelized to use multithreading, so we configure a maximum of 12
threads for all experiments.  Our experiments are run on a computer
with an AMD Ryzen Threadripper 2950X 16-core processor, with a
30-minute timeout per differential verification task.

While we could try and adapt a single-network verification tool to our
task as done previously~\cite{PaulsenWW20}, we note that \ReluDiff{}
has been shown to significantly outperform (by several orders of
magnitude) this naive approach.

\subsection{Results}

In the remainder of this section, we present our experimental results
in two steps.  First, we present the overall verification results on
all benchmarks. Then, we focus on the detailed verification results on
the more difficult verification tasks.

\subsubsection{Summary of Results on All Benchmarks}

Our experimental results show that, on all benchmarks, the
improved \ReluDiffP{} slightly but consistently outperforms the
original \ReluDiff{} due to its use of the more accurate component
from \Neurify{} instead of \ReluVal{} for bounding the absolute values
of individual neurons.  Thus, to save space, we will only show the
results that compare \Name{} (our method) and \ReluDiffP{}.

\begin{figure}
\centering
\includegraphics[width=0.7\linewidth]{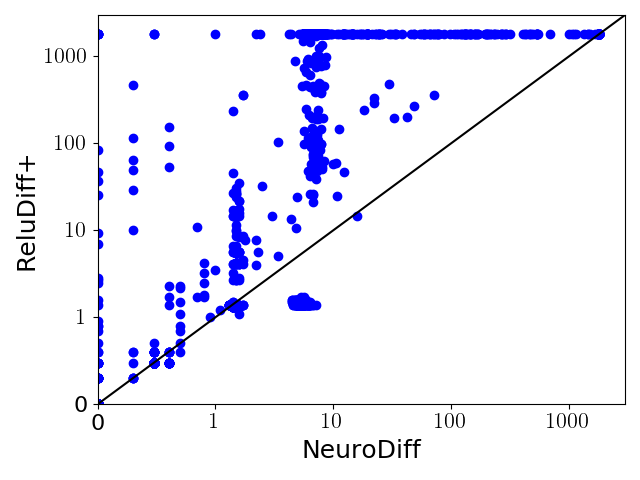}
\caption{Comparing the execution times of \Name{} and \ReluDiffP{} on all verification tasks.}
\label{fig:expsummary}
\end{figure}

We summarize the comparison between \Name{} and \ReluDiffP{} using a
scatter plot in Figure~\ref{fig:expsummary}, where each point
represents a differential verification task: the x-axis is the
execution time of \Name{} in seconds, and the y-axis the execution
time of \ReluDiffP{} in seconds.  Thus, points on the diagonal line
are ties, while points above the diagonal line are wins for \Name{}.

The results show that \Name{} outperformed \ReluDiffP{} for most
verification tasks.  Since the execution time is in logrithmic scale
 the speedups of \Name{} are more
than 1000X for many of these verification tasks.
While there are cases where \Name{} is slower than \ReluDiffP{}, due
to the overhead of adding symbolic variables, the differences are on the
order of seconds.  Since they are all on the small MNIST networks and the
HAR network that are very easy for both tools, we omit
an in-depth analysis of them.

In the remainder of this section, we present an in-depth analysis of
the more difficult verification tasks.

\subsubsection{Results on ACAS Networks}

\begin{table}
\caption{Results for ACAS networks with $ \epsilon = 0.05 $.}
\label{tbl:acas_orig}
	\scalebox{0.75}{
		\begin{tabular}{|c|ccc|ccc|c|}\hline
			\multirow{2}{*}{\makecell[c]{Property}} &  \multicolumn{3}{c|}{\Name{} (new)}
			& \multicolumn{3}{c|}{\ReluDiffP{}} & \multirow{2}{*}{Speedup} \\
			\cline{2-7}
			& proved  & undet. & time (s) & proved & undet. & time (s) &  \\\hline
			\hline

        $\varphi_{1}$   & 45    & 0     & 522.6         & 44    & 1     & 4800.6        &   9.2         \\\hline
$\varphi_{3}$   & 42    & 0     &   2.3         & 42    & 0     &   4.1         &   1.8         \\\hline
$\varphi_{4}$   & 42    & 0     &   1.7         & 42    & 0     &   2.8         &   1.7         \\\hline
$\varphi_{5}$   & 1     & 0     &   0.2         & 1     & 0     &   0.2         &   1.4         \\\hline
$\varphi_{6}$  & 2     & 0     &   0.6         & 2     & 0     &   0.4         &   0.7         \\\hline
$\varphi_{7}$   & 1     & 0     & 1404.4        & 0     & 1     & 1800.0        &   1.3         \\\hline
$\varphi_{8}$   & 1     & 0     & 132.2         & 1     & 0     & 361.8         &   2.7         \\\hline
$\varphi_{9}$   & 1     & 0     &   0.6         & 1     & 0     &   2.3         &   3.7         \\\hline
$\varphi_{10}$  & 1     & 0     &   0.9         & 1     & 0     &   0.7         &   0.8         \\\hline
$\varphi_{11}$  & 1     & 0     &   0.2         & 1     & 0     &   0.3         &   1.6         \\\hline
$\varphi_{12}$  & 1     & 0     &   2.8         & 1     & 0     & 360.9         & 129.4         \\\hline
$\varphi_{13}$  & 1     & 0     &   5.8         & 1     & 0     &   5.1         &   0.9         \\\hline
$\varphi_{14}$  & 2     & 0     &   0.5         & 2     & 0     &  95.9         & 196.2         \\\hline
$\varphi_{15}$  & 2     & 0     &   0.6         & 2     & 0     &  65.0         & 113.2         \\\hline

		\end{tabular}
	}
\end{table}

\begin{table}
\caption{Results for ACAS networks with $ \epsilon = 0.01 $.}
\label{tbl:acas_new}
	\scalebox{0.75}{
		\begin{tabular}{|c|ccc|ccc|c|}\hline
			\multirow{2}{*}{\makecell[c]{Property}} &  \multicolumn{3}{c|}{\Name{} (new)}
			& \multicolumn{3}{c|}{\ReluDiffP{}} & \multirow{2}{*}{Speedup} \\
			\cline{2-7}
			& proved  & undet. & time (s) & proved & undet. & time (s) &  \\\hline
			\hline

        $\varphi_{1}$   & 41    & 4     & 11400.1       & 15    & 30    & 55778.6       &   4.9         \\\hline
$\varphi_{3}$   & 42    & 0     &  14.3         & 35    & 7     & 13642.2       & 957.2         \\\hline
$\varphi_{4}$   & 42    & 0     &   3.8         & 37    & 5     & 9115.0        & 2390.1        \\\hline
$\varphi_{5}$   & 1     & 0     &   0.3         & 0     & 1     & 1800.0        & 5520.5        \\\hline
$\varphi_{16}$  & 2     & 0     &   1.0         & 2     & 0     &   0.8         &   0.8         \\\hline
$\varphi_{7}$   & 0     & 1     & 1800.0        & 0     & 1     & 1800.0        &   1.0         \\\hline
$\varphi_{8}$   & 1     & 0     & 1115.9        & 0     & 1     & 1800.0        &   1.6         \\\hline
$\varphi_{9}$   & 1     & 0     &   2.4         & 0     & 1     & 1800.0        & 738.2         \\\hline
$\varphi_{10}$  & 1     & 0     &   1.6         & 1     & 0     &   1.1         &   0.7         \\\hline
$\varphi_{11}$  & 1     & 0     &   0.3         & 0     & 1     & 1800.0        & 5673.8        \\\hline
$\varphi_{12}$  & 1     & 0     & 132.2         & 0     & 1     & 1800.0        &  13.6         \\\hline
$\varphi_{13}$  & 1     & 0     &  15.9         & 1     & 0     &  14.8         &   0.9         \\\hline
$\varphi_{14}$  & 2     & 0     & 1589.3        & 0     & 2     & 3600.0        &   2.3         \\\hline
$\varphi_{15}$  & 2     & 0     & 579.4         & 0     & 2     & 3600.0        &   6.2         \\\hline

		\end{tabular}
	}
\end{table}

For ACAS networks, we consider two different sets of properties,
namely the original properties from Paulsen et al.~\cite{PaulsenWW20}
where $ \epsilon = 0.05 $, and the same properties but with $ \epsilon
= 0.01 $.  We emphasize that, while verifying $ \epsilon = 0.05 $ is
useful, this means that the output value can vary by up to
10\%. Considering $ \epsilon = 0.01 $ means that the output value can
vary by up to 2\%, which is much more useful.
%For all properties, we
%let \Name{} use up to 15 intermediate variables. We tuned the number
%on property $ \varphi_1 $ on the first of the 45 networks
%(i.e. network 1\_1).

Our results are shown in Tables~\ref{tbl:acas_orig}
and~\ref{tbl:acas_new}, where the first column shows the property,
which defines the input space considered. The next three columns show
the results for \Name{}, specifically the number of verified networks
(out of the 45 networks), the number of unverified networks, and the
total run time across all networks. The next three show the same
results, but for \ReluDiffP{}. The final column shows the average
speed up of \Name{}.

The results show that \Name{} makes significant gains in both speed
and accuracy. Specifically, the speedups are up to two and three
orders of magnitude for $ \epsilon = 0.05 $ and $ 0.01 $,
respectively. In addition, at the more accurate $ \epsilon = 0.01 $
level, \Name{} is able to complete 53 more verification tasks, out of
the total 142 verification tasks.

\subsubsection{Results on MNIST Networks}

\begin{figure*}[t]
\centering
	\begin{minipage}[t]{0.498\linewidth}
        \centering
		\captionsetup{width=0.9\textwidth}
		\includegraphics[width=.7\linewidth]{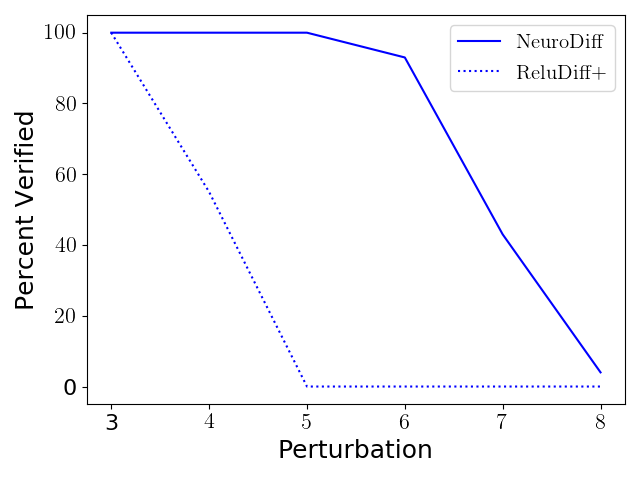}
		\caption{Percentage of verification tasks completed on the MNIST 4x1024 network for various perturbations.\label{fig:perturbation_comp}}
	\end{minipage}
	\begin{minipage}[t]{0.498\linewidth}
        \centering
		\captionsetup{width=0.9\textwidth}
		\includegraphics[width=.7\linewidth]{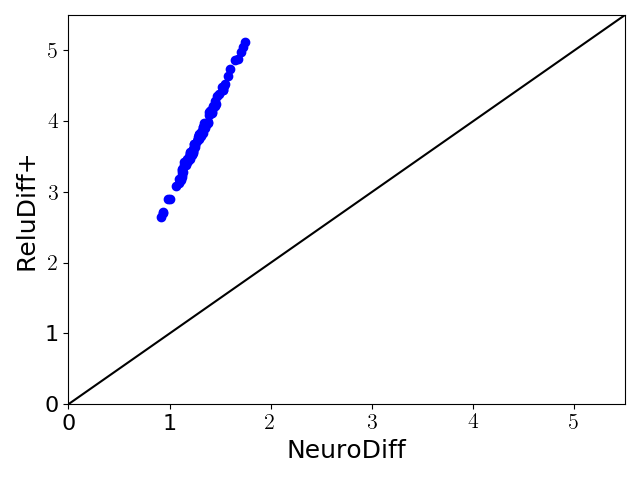}
		\caption{Accuracy comparison for a single forward pass on the MNIST 4x1024 network with perturbation of 8.\label{fig:accuracy_comp}}
	\end{minipage}%
\end{figure*}

For MNIST, we focus on the 4x1024 network, which is the largest
network considered by Paulsen et al.~\cite{PaulsenWW20}.  In contrast,
since the smaller networks, namely 3x100 and 2x512 networks, were
handled easily by both tools, we omit their results.
In the MNIST-related verification tasks, the goal is to verify
$ \epsilon = 1 $ for the given input region. We consider the two types
of input regions from the previous work, namely global perturbations
and targeted pixel perturbations, however we use input regions that
are hundreds of orders of magnitude larger.
%For all verification
%tasks, we allow 1,000 intermediate variables for \Name{}. We tuned the
%number on the first of our global perturbation properties.

First, we look at the global perturbation. For these, the input space
is created by taking an input image and then allowing a perturbation
of +/- $ p $ greyscale units to all of its pixels. In the previous
work, the largest perturbation was $ p = 3$.
Figure~\ref{fig:perturbation_comp} compares \Name{} and \ReluDiffP{}
on $ p = 3 $ all the way up to 8, where the x-axis is the perturbation
applied, and the y-axis is the percentage of verification tasks (out
of 100) that each can handle.

The results show that \Name{} can handle perturbations up to +/-
6 units, whereas \ReluDiffP{} begins to struggle at 4. While the
difference between 4 and 6, may seem small, the volume of input space
for a perturbation of 6 is $ 6^{784}/4^{784} \approx 1.1 \times
10^{138} $ times larger than 4, or in other words, 138 orders of
magnitude larger.

Next, we show a comparison of the epsilon verified by a single forward pass
for a perturbation of 8 on the MNIST 4x1024 network in Figure~\ref{fig:accuracy_comp}.
Points above the blue line indicate \Name{} performed
better. Overall, \Name{} is between two and three times more accurate
than \ReluDiffP{}.

Finally, we look at the targeted pixel perturbation properties. For
these, the input space is created by taking an image, randomly
choosing $ n $ pixels, and setting there bounds to $ [0,255] $, i.e.,
allowing arbitrary changes to the chosen pixels. We again use the
4x1024 MNIST network. The results are summarized in
Table~\ref{tbl:MNIST_pix}. The first column shows the number of
randomly perturbed pixels. We can again see very large speedups, and a
significant increase in the size of the input region that \Name{} can
handle.

\begin{table}
	\caption{Results of the MNIST 4x1024  pixel experiment.}
	\label{tbl:MNIST_pix}
	\scalebox{0.75}{
		\begin{tabular}{|c|ccc|ccc|c|}\hline
			\multirow{2}{*}{\makecell[c]{Num. \\ Pixels}} &  \multicolumn{3}{c|}{\Name{} (new)}
			& \multicolumn{3}{c|}{\ReluDiffP{}} & \multirow{2}{*}{Speedup} \\
			\cline{2-7}
			& proved  & undet. & time (s) & proved & undet. & time (s) &  \\\hline
			\hline

15      & 100   & 0     & 236.5 & 100   & 0     & 1610.2        &   6.8 \\\hline
18      & 100   & 0     & 540.8 & 88    & 12    & 34505.8       &  63.8 \\\hline
21      & 100   & 0     & 1004.0        & 30    & 70    & 145064.5      & 144.5 \\\hline
24      & 99    & 1     & 7860.1        & 1     & 99    & 179715.9      &  22.9 \\\hline
27      & 83    & 17    & 49824.0       & 0     & 100   & 180000.0      &   3.6 \\\hline

		\end{tabular}
	}
\end{table}

\subsubsection{Contribution of Each Technique}

Here, we analyze the contribution of individual techniques, namely
convex approximations and symbolic variables, to the overall
performance improvement.

In Table~\ref{tbl:MNIST_comp}, we present the average $ \epsilon $
that was able to be verified after a single forward pass on the
4x1024 MNIST network for each of
the four techniques: \ReluDiffP{} (baseline), \Name{} with only convex
approximations, \Name{} with only intermediate variables, and the
full \Name{}.

Overall, the individual benefits of the two proposed approximation
techniques are obvious.  While convex approximation
(alone) consistently provides benefit as perturbation increases, the
benefit of symbolic variables (alone) tends to decrease.
In addition, combining the two provides much greater benefit than the
sum of their individual contributions. With perturbation of 8, for
example, convex approximations alone are 1.59 times more accurate than
\ReluDiffP{}, and intermediate variables alone are 1.01 times more accurate.
However, together they are 2.93 times more accurate.

\begin{table}
	\caption{Evaluating the individual contributions of convex approximation and symbolic variables using the  MNIST 4x1024 global perturbation experiment.}
	\label{tbl:MNIST_comp}
	\scalebox{0.75}{
		\begin{tabular}{|c|c|c|c|c|}\hline
			\multirow{2}{*}{Perturb} &  \multicolumn{4}{c|}{Average $ \epsilon $ Verified} \\
			\cline{2-5}
			& \ReluDiffP{} & Conv. Approx. & Int. Vars. & \Name{} \\
			\hline
3 & 0.59 & 0.42 (+1.39x) & 0.43 (+1.38x) & 0.20 (+2.93x) \\\hline
4 & 1.02 & 0.70 (+1.46x) & 0.87 (+1.18x) & 0.36 (+2.85x) \\\hline
5 & 1.60 & 1.06 (+1.52x) & 1.47 (+1.09x) & 0.56 (+2.87x) \\\hline
6 & 2.29 & 1.47 (+1.55x) & 2.19 (+1.04x) & 0.79 (+2.90x) \\\hline
7 & 3.02 & 1.92 (+1.58x) & 2.96 (+1.02x) & 1.04 (+2.91x) \\\hline
8 & 3.80 & 2.39 (+1.59x) & 3.77 (+1.01x) & 1.30 (+2.93x) \\\hline
		\end{tabular}
	}
\end{table}

The results suggest two things.
First, intermediate
symbolic variables perform well when a significant portion of the
network is already in the stable state.  We confirm, by manually
inspecting the experimental results, that it is indeed the case when
we use a perturbation of 3 and 8 in the MNIST experiments.
%, and we believe
%this is the case for ACAS as well, once the input space is
%sufficiently partitioned.
Second, the convex approximations provide the most benefit when the
pre-ReLU delta intervals are (1) significantly wide, and (2) still
contain a significant amount of symbolic information. This is also
confirmed by manually inspecting our MNIST results: increasing the
perturbation increases the overall width of the delta intervals.
%In
%addition, this is very likely why we do not see a significant benefit
%from convex approximations in the ACAS results. Since the ACAS
%networks have many fewer parameters than the MNIST network, the total
%amount of change in the network from rounding to 16 bits is relatively
%small. This leads to relatively narrow delta intervals, and hence
%there is little opportunity for benefit.

%\subsection{Threats to Validity}
%We note that the refinement of both \Name{} and \ReluDiff{} typically does not work well with large input spaces. However, the LP solving-based refinement of Wang et al.~\cite{WangPWYJ18nips} has been shown to scale better to higher-dimensional input spaces, and could straightforwardly be integrated into \Name{}.
%
%We also focus on feed-forward ReLU networks. ReLU activations are preferable in safety critical systems because they are the most amenable to verification. Still, much previous work~\cite{SinghGPV19, SinghGPV19iclr, zhang2018efficient} has shown that convex approximations can be extended non-linear activations as well. We believe \Name{} can be extended as such too.

\section{Related Work}
\label{sec:related}

% Single network verification
% - complete techniques
Aside from \ReluDiff{}~\cite{PaulsenWW20}, the most closely related to
our work are those that focus on verifying properties of single
networks as opposed to two or more networks. These verification
approaches can be broadly categorized into those that use exact,
constraint solving-based techniques and those that use approximations.

On the constraint solving side, several works have adapted
off-the-shelf solvers~\cite{CarliniW17, tjeng2019evaluating,
BastaniILVNC16, Ehlers17, baluta2019quantitative}, or even implemented
solvers specifically for neural networks~\cite{KatzBDJK17,
KatzHIJLLSTWZDK19}.
On the approximation side, many use techniques that fit into the framework of abstract interpretation~\cite{CousotC77}. For example, many works have leveraged abstract domains such as intervals~\cite{WangPWYJ18, WengZCSHDBD18, zhang2018efficient, JulianKO18}, polyhedra~\cite{Singh2019krelu, SinghGPV19}, and zonotopes~\cite{SinghGPV19iclr, GehrMDTCV18}.

In addition, these verification techniques have also been
combined~\cite{SinghGPV19iclr, WangPWYJ18nips, HuangKWW17}, or
entirely different approaches~\cite{RuanHK18, DvijothamSGMK18,
GopinathKPB18}, such as bounding a network's lipschitz constant, have
been studied. These verification techniques can also be integrated
into the training process to produce more robust and easier to verify
networks~\cite{FischerBDGZV19, MirmanGV18, WongK18,
balunovic2020adversarial}. These works are orthogonal, though we
believe their techniques can be adapted to our domain.

% Various heuristic based approaches, focus on discovering but not proving the absence of adversarial examples
A related but tangential line of work focuses on discovering
interesting behaviors of neural networks, though without any
guarantees.
Most closely related to our work are differential testing
techniques~\cite{xie2019diffchaser, PeiCYJ17, MaLLZG18}, which focus
on finding disagreements between a set of networks.
However, these techniques do not attempt to prove the equivalence or
similarity of multiple networks.

Other works are more geared towards single network testing, and use
white-box testing techniques~\cite{ma2018deepgauge, xie2019deephunter,
SunWRHKK18, TianPJR18, odena2018tensorfuzz}, such as neuron coverage
statistics, to assess how well a network has been tested, and also
report interesting behaviors. Both of these can be thought of as
adapting software engineering techniques to machine learning.

In addition, many works use machine learning techniques, such as
gradient optimization, to find interesting behaviors, such as
adversarial examples\cite{KurakinGB17a, madry2017towards, NguyenYC15,
XuQE16, Moosavi-Dezfooli16}. These interesting behaviors can then be
used to retrain the network to improve
robustness~\cite{GoodfellowSS15, RaghunathanSL18}. Again, these
techniques do not provide guarantees, though we believe they could be
integrated into \Name{} to quickly find counterexamples.

Finally, our work draws inspiration from classic software engineering
techniques, such as regression testing~\cite{rothermel1997safe},
differential assertion checking~\cite{lahiri2013differential},
differential fuzzing~\cite{NilizadehNP19}, and incremental symbolic
execution~\cite{Person11,GuoKW16}, where one version of a program is
used as an ``oracle'', to more efficiently test or verify a new
version of the same program.  In our case, $ f $ can be thought of as
the oracle, while $ f' $ is the new version.

\section{Conclusions}
\label{sec:conclusion}

We have presented \Name{}, a scalable differential verification
technique for soundly bounding the difference between two feed-forward
neural networks. \Name{} leverages novel convex approximations, which
reduce the overall approximation error, and intermediate symbolic
variables, which control the error explosion, to significantly improve
efficiency and accuracy of the analysis. Our experimental evaluation
shows that \Name{} can achieve up to 1000X speedup and is up to five
times as accurate.

\section*{Acknowledgments}

This work was partially funded by the U.S. Office of Naval Research (ONR) under
the grant N00014-17-1-2896.

\newpage
\clearpage
\bibliography{main}
\end{document}